\documentclass{article}

\usepackage{microtype}
\usepackage{graphicx}
\usepackage{subcaption}
\usepackage{booktabs} % for professional tables

\usepackage{hyperref}

\usepackage[preprint]{icml2026}

\usepackage{amsmath}
\usepackage{amssymb}
\usepackage{mathtools}
\usepackage{amsthm}

\usepackage[capitalize,noabbrev]{cleveref}

\theoremstyle{plain}

\theoremstyle{definition}

\theoremstyle{remark}

\usepackage[textsize=tiny]{todonotes}

\icmltitlerunning{Supervised sparse auto-encoders for interpretable and compositional representations}

\begin{document}

\twocolumn[
  \icmltitle{Supervised sparse auto-encoders for interpretable and compositional representations}



  \icmlsetsymbol{equal}{*}

  \begin{icmlauthorlist}
    \icmlauthor{Ouns El Harzli}{equal,1}
    \icmlauthor{Hugo Wallner}{equal,2}
    \icmlauthor{Yoonsoo Nam}{3}
    \icmlauthor{Haixuan Xavier Tao}{4,5}
    %\icmlauthor{Firstname5 Lastname5}{yyy}
    %\icmlauthor{Firstname6 Lastname6}{sch,yyy,comp}
    %\icmlauthor{Firstname7 Lastname7}{comp}
    %\icmlauthor{Firstname8 Lastname8}{sch}
    %\icmlauthor{Firstname8 Lastname8}{yyy,comp}
  \end{icmlauthorlist}

  \icmlaffiliation{1}{Department of Computer Science, University of Oxford, Oxford, UK}
  \icmlaffiliation{3}{KAIST AI, Korean Advanced Institute of Science and Technology, Seoul, South Korea}
  \icmlaffiliation{2}{Independent researcher}
  \icmlaffiliation{4}{1ms.ai, Paris, France}
  \icmlaffiliation{5}{Hugging Face, Paris, France}
  %\icmlaffiliation{comp}{Company Name, Location, Country}
  %\icmlaffiliation{sch}{School of ZZZ, Institute of WWW, Location, Country}

  \icmlcorrespondingauthor{Ouns El Harzli}{ouns.elharzli@new.ox.ac.uk}
  \icmlcorrespondingauthor{Hugo Wallner}{hugo.p.wallner@gmail.com}

  \icmlkeywords{Machine Learning, ICML}

  \vskip 0.3in
]



%\printAffiliationsAndNotice{}  % no special notice (required even if empty)
\printAffiliationsAndNotice{\icmlEqualContribution}

\begin{abstract}
Sparse auto-encoders (SAEs) have re-emerged as a prominent method for mechanistic interpretability, yet they face two significant challenges: the non-smoothness of the $L_1$ penalty, which hinders reconstruction and scalability, and a lack of alignment between learned features and human semantics. In this paper, we address these limitations by adapting unconstrained feature models--a mathematical framework from neural collapse theory--and by supervising the task. We supervise (decoder-only) SAEs to reconstruct feature vectors by jointly learning sparse concept embeddings and decoder weights. Validated on Stable Diffusion 3.5, our approach demonstrates compositional generalization, successfully reconstructing images with concept combinations unseen during training, and enabling feature-level intervention for semantic image editing without prompt modification. 
\end{abstract}

\section{Introduction}

The advent of large foundation models in language, vision, and multimodal domains has renewed interest in methods for gaining interpretability and control over learned representations. In particular, recent work on \textit{sparse autoencoders} (SAEs) has demonstrated their ability to identify selective and localized features within large-scale models \cite{bricken2023towards, kissane2024interpreting, surkov2025one, tian2025sparse}. These sparse representations are promising tools for mechanistic interpretability, especially for discovering features in internal activations or embeddings that are causally relevant to model behavior.

Despite this promise, SAEs trained in an \textit{unsupervised} fashion suffer from key limitations: they often learn features that are entangled, noisy, or uninterpretable, i.e.\ misaligned with human-understandable semantic concepts \cite{smith2025negative}. Moreover, imposing sparsity in an unsupervised setting is achieved by training with $L_1$ regularization, which introduces optimization challenges that grow with dimensionality \cite{ng2011sparseae}, leading to instability and poor reconstruction fidelity \cite{gao2025scaling}. These challenges limit the scalability and the applicability of SAEs to interpretability in large models \cite{smith2025negative}.

In an orthogonal (theoretical) research thread, the \emph{unconstrained feature model} has been introduced as a proxy to neural networks, sharing their training dynamics and provably reproducing the neural collapse phenomenon \cite{suken2023deep}. In this theoretical framework, features are treated as free parameters, i.e.\ there are not tied to any input data \cite{e2022emergence,tirer2022extended}, and are only trained with respect to the outputs. This last property has limited the practicality of unconstrained feature model, since it cannot learn from the input data, and up until now, unconstrained feature models have been confined to theoretical study of the dynamics of gradient-based training. Our first insight was to notice that (sparse) auto-encoding is well suited for applying this framework, since the input and the ouput data are intended to coincide, hence we are not missing any information by ignoring the inputs, and training solely with respect to the outputs is sound.

%In spirit, our approach reflects a similar design: we treat the prompt embedding as an editable feature vector and optimize a fixed decoder to map sparse concept codes to target embeddings. However, we impose structure in the form of sparsity and concept alignment, rather than learning arbitrary features. While the UFM assumes complete flexibility in internal representations, our model restricts edits to a meaningful, compositional basis—trading full generality for interpretability and control. In this sense, our method bridges ideas from UFM-style linear analysis with structured latent representations suitable for editing in large generative models.

Based on the unconstrained feature model, we propose in this work a \emph{supervised sparse auto-encoder} (SSAE) framework that resolves by design several of the challenges faced by unsupervised SAEs. Instead of discovering sparse features as part of the training of the SAE , we \textit{define} them upfront through \emph{sparse feature design}: our method constructs a sparse latent space aligned with a known concept dictionary, and the non-zero coefficients play the role of the free features of an unconstrained feature model. This allows us to learn a decoder-only model to reconstruct the actual feature space from these sparse vectors, where the sparse structure readily defines which concepts appear in which feature vector. This framework removes the need for a trained encoder \footnote{Our framework does not \emph{exclude} the use of an encoder: it can be incorporated as explained in Paragraph \ref{paragraph-encoders}, but it is not needed.}, avoids $L_1$ penalties entirely, and guarantees interpretability through structure. In our SSAE framework, each concept in the dictionary is associated with a sub-vector in the sparse latent space. This framework thus supports \emph{compositional generalisation}, i.e.\ combining concepts that were never seen together in the training set (by imputing the learned values to the corresponding sub-vectors), and reconstructing a new feature vector which contains the semantics of both concepts. Using theoretical arguments from the unconstrained feature model literature, we argue that our framework actually \emph{encourages} compositional generalisation, by promoting decorrelation between concept subspaces.

To verify the soundness of our framework, we apply our methodology to the \emph{prompt embedding space} of Stable Diffusion 3.5. We construct sparse concept codes corresponding to interpretable attributes (e.g., \texttt{blond hair}, \texttt{gun}, \texttt{standing}), and to demonstrate compositional generalisation, we show that our SSAE model is able to reconstruct images containing concepts that were not seen together in its training set. We present preliminary experimental results showcasing that this methodlogy can be used to edit prompt embeddings in a modular fashion---removing, adding, or replacing individual attributes---without modifying the original prompt text. The result is image-level editing through direct semantic manipulation of feature space.

Our method is simple, task- and model-agnostic: while we empirically explored only the application to prompt embeddings for Stable Diffusion 3.5, the same framework could be applied to hidden activations in transformers, U-Net layers, or other hidden representations in foundation models. We view this as a step toward interpretable, structured interfaces to large models, where sparse supervision replaces unsupervised discovery, and we hope this work will motivate many applications of our methodology.

\textbf{Contributions.} Our main contributions are:
\begin{itemize}
\item We introduce a (decoder-only) supervised SAE framework using predefined sparse concept structure and supporting decoder-only training, addressing by design two of the main limitations of unsupersived SAEs (namely, $L_1$ penalty and semantic alignment).
\item We showcase that our methodology supports compositional generalisation on prompt embeddings in Stable Diffusion 3.5.
\item We explore a first application of our methodology to modular editing via feature-level intervention.
\end{itemize}

\begin{figure*}[h!]
    \centering
        \centering
        \includegraphics[width=\textwidth]{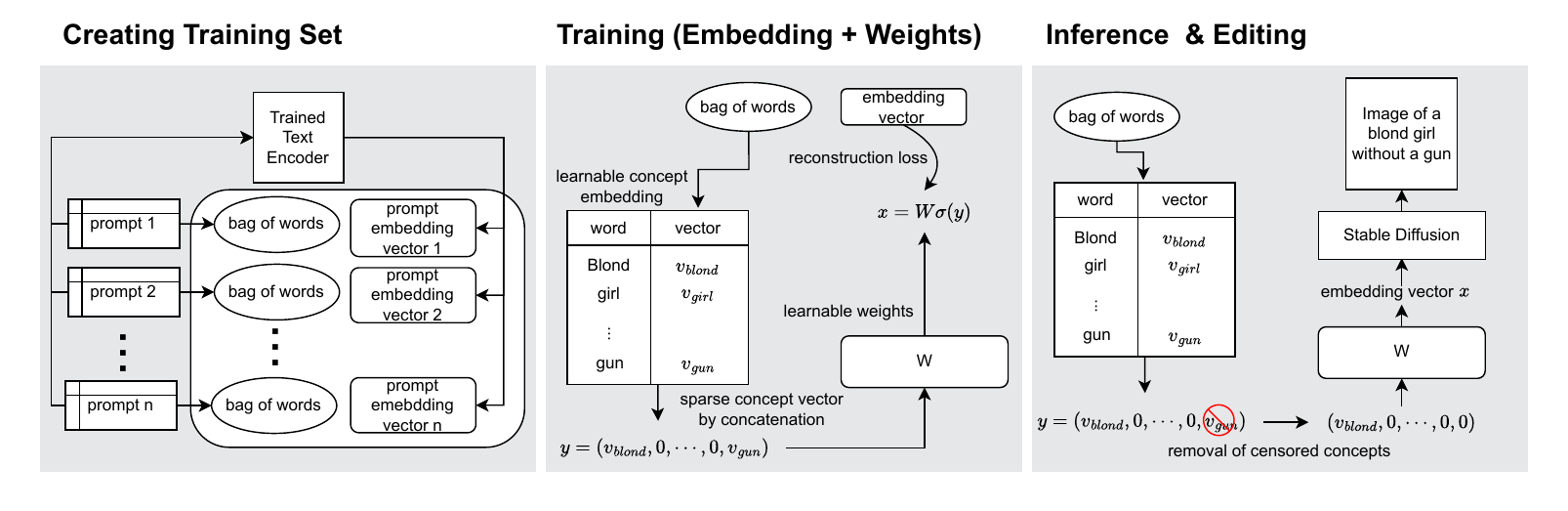}
    \caption{Editing workflow: We train weights of a decoder-only SSAE $\mathbf{W}_2$ and the dictionary of concepts (i.e. trainable parameters of $\mathbf{Y}$) to reconstruct prompt embeddings for image generation. We use the trained dictionary (sparse latent space) and $\mathbf{W}_2$ to add/modify/remove to guarantee safer image generation at inference time.}
    \label{fig:money}
\end{figure*}

\section{Preliminaries}

\paragraph{Sparse auto-encoders.}

An auto-encoder is a neural network designed to learn a compressed representation of input data in an unsupervised manner. Given input $\mathbf{x} \in \mathbb{R}^d$, the network consists of two components: an \emph{encoder} function $f_{\theta}: \mathbb{R}^d \to \mathbb{R}^m$ that maps the input to a latent representation $\mathbf{h} = f_{\theta}(\mathbf{x})$, and a \emph{decoder} function $g_{\phi}: \mathbb{R}^m \to \mathbb{R}^d$ that reconstructs the input as $\hat{\mathbf{x}} = g_{\phi}(\mathbf{h})$.

A \emph{sparse auto-encoder} encourages the hidden representation $\mathbf{h}$ to be sparse, i.e., most of its components are close to zero. This is typically achieved by adding a regularization term to the loss function that penalizes non-sparse activations. One common approach is to use an $L_1$ penalty on the hidden units, yielding the following objective:
\begin{equation}
    \mathcal{L} = \frac{1}{n} \sum_{i=1}^n \left\| \mathbf{x}^{(i)} - g_{\phi}(f_{\theta}(\mathbf{x}^{(i)})) \right\|_2^2 + \lambda \left\| f_{\theta}(\mathbf{x}^{(i)}) \right\|_1
\end{equation}
where $\lambda > 0$ is a sparsity regularization coefficient, and \( n \) is the number of training examples.

This formulation encourages each hidden representation \( \mathbf{h}^{(i)} \) to be sparse, thus promoting feature selectivity and interpretability in the learned representations.

\paragraph{Unconstrained feature model.}
The unconstrained feature model is a theoretical framework used to analyze learning with free features~\cite{tirer2022extended,e2022emergence}. Given input-label pairs $(x_i, y_i)$, instead of assuming that the features $\phi(\mathbf{x}_i)$ are outputs of a known neural network, the unconstrained feature model treats them as free parameters $\mathbf{z}_i \in \mathbb{R}^d$ to be optimized. The model minimizes a supervised loss over both feature vectors $\{\mathbf{z}_i\}_{i=1}^n$ and a final linear map $\mathbf{w} \in \mathbb{R}^d$:
\begin{equation}
    \min_{\{\mathbf{z}_i\}, \mathbf{w}} \quad \sum_{i=1}^n \ell(y_i, \mathbf{w}^\top \sigma (\mathbf{z}_i)) + \lambda R(\{\mathbf{z}_i\}),
\end{equation}
where $\ell$ is a loss function (e.g., cross-entropy), $\sigma$ is an activation function (e.g.\ ReLU) and $R$ is a regularizer (often $\ell_2$ norm or a centroid constraint). The unconstrained feature model allows one to analyze generalization and feature structure in isolation from data and architectural assumptions. 
%In our case, although the features (concept codes) are not arbitrary—being structured and sparse—we similarly fix a decoder $W$ and treat the latent inputs $y_i$ as tunable codes aligned with interpretable concepts.

\paragraph{Compositional generalization.}
Compositional generalization refers to a model's ability to generalize to novel combinations of known elements~\cite{lake2018generalization}. Formally, suppose we define a concept space $\mathcal{C}$, where each concept $c_k \in \mathcal{C}$ is associated with a latent vector $v_k \in \mathbb{R}^d$. A compositional representation treats inputs as structured combinations of such concepts. For instance, a composite latent can be formed by summing active components:
\begin{equation}
    z = \sum_{k \in S} v_k, \quad S \subseteq \{1, \dots, K\},
\end{equation}
where $S$ is the set of active concepts in an example. A model exhibits compositional generalization if it can correctly process or generate outputs for unseen combinations $S'$ not present in training.

\section{Methodology}

\paragraph{Training set.} Consider $\mathcal{C} =\{c_1, ..., c_K\}$ a set of concepts that we want to isolate in a particular feature map or a prompt embedding $\phi (\cdot)$. 

For example, for a text-to-image generation task: \begin{align*}
    \mathcal{C} = & \{ \texttt{blue eyes}, \texttt{blond hair}, \texttt{brune hair}, \\& \texttt{black eyes}, 
    \texttt{being seated}, \\& \texttt{standing in the street}, \\& \texttt{being on horseback}, ... \}~.
\end{align*} 

Construct $n$ realisations that contain different subsets of these concepts: we note $S_i \subset \mathcal{C}$ for $i \in \{1, ...,n\}$ these subsets. For text-to-image generation, we start from $n$ prompts that contains the concepts and we propagate to the feature map or to the prompt embedding of interest. We obtain a training data matrix $\mathbf{X}^{N \times n}$, where $N$ is the dimensionality of the feature map or prompt embedding $\phi (\cdot)$. $\mathbf{X}$ is the matrix we are trying to reconstruct with our SSAE.

\paragraph{Sparse feature design.}

The sparse feature matrix is defined as follows:
\begin{align*}
    \mathbf{Y} \in \mathbb{R}^{(d \cdot K) \times n}
\end{align*}
where $d$ is an hyperparameter that corresponds to the dimensionality of the subspace needed to encode each concept, and verifies for each $i \in \{1, ..., n\}$,  $j \in \{1, ..., d\}$, and $k \in \{1, ..., K\}$, $\mathbf{Y}_{j \cdot k, i} =0$ if $c_k \notin S_i$, and $\mathbf{Y}_{j \cdot k, i} = y_{j,k}$ is a trainable parameter if $c_k \in S_i$. This design ensures that for each $i \in \{1, ..., n\}$, the sub-vectors associated to a concept $c_k$ which is not in $S_i$ are set to zero, whereas the components associated to a concept $c_k$ which is in $S_i$ are learnable. Importantly, the learnable parameters $y_{j,k}$ are set to not depend on $i$, i.e.\ the SSAE will learn a single representation for each concept $c_k$ across all realisations $i \in \{1, ..., n\}$: the sub-vector $(y_{j,k})_{j \in \{1, ..., d\}}$.

\paragraph{Decoder-only SSAE training.} Training a \emph{decoder-only} supervised sparse auto-encoder (SSAE) to reconstruct $\mathbf{X}$ consists in minimizing the loss:
\begin{align*}
    \mathcal{L} = \mathbin\Vert \mathbf{X} - \mathbf{W}_2 \sigma (\mathbf{Y}) \mathbin\Vert_2
\end{align*}

where $\sigma$ is an activation function (e.g.\ ReLU) and $\mathbf{W}_2 \in \mathbb{R}^{N \times (d \cdot K)}$ is the decoder matrix. This formulation is exactly an unconstrained feature model, and is fully differentiable with respect to $\mathbf{W}_2$ and $\mathbf{Y}$. Unlike unsupervised SAEs, there is no $L_1$ penalty on $\mathbf{Y}$, and no encoder $f_\theta$ is learned: sparsity is predefined to align with semantic concepts. 

Training can be batched with respect to the number of inputs $n$, thus allowing for efficient gradient-based training. Gradient updates scale linearly with the number of concepts $K$ and the concept subspace dimension $d$.

\paragraph{Assessing compositional generalization.} In this paragraph, we detail how our decoder-only SSAE readily supports compositional generalization. We use \emph{semantic composition} to verify that the concepts have correctly been isolated and generalize in terms of semantics. Consider two concepts $c_{k_1}, c_{k_2}$ that do not appear together in the training set, i.e.\ there is no $i \in \{1, ...,n\}$ such that $c_{k_1} \in S_i$ and $c_{k_2} \in S_i$. Starting from any column of the trained $\mathbf{Y}$, we can construct a new sparse feature vector $\mathbf{y}$ in the same sparse latent space as $\mathbf{Y}$, by assigning the learned values $(y_{j,{k_1}})_{j \in \{1, ..., d\}}$ and $(y_{j,{k_2}})_{j \in \{1, ..., d\}}$ to the corresponding components (essentially, concatenating sub-vectors). Applying the trained decoder, we obtain $\mathbf{W}_2 \sigma (\mathbf{y}) \in \mathbb{R}^{N}$ which is a new representation at the feature map or prompt embedding level (i.e.\ in the feature space that we are trying to reconstruct). One can then propagate this new representation in the rest of the architecture; for example applying Stable Diffusion on a new reconstructed prompt embedding, or propagating the rest of a U-net or Transformer architecture on a new reconstructed feature map; and inspect whether the final output does contain concepts $c_1$ and $c_2$, and whether they interfere with other concepts. Compositional generalisation is a property that enables interpretability: one can explore which concepts are entangled with one another, and the causal relation of different concepts on the outputs.

As a direct consequence of decoder-only SSAE being an unconstrained constrained model, we expect implicit bias and decorrelation of concept subspace to emerge from gradient-based training. Indeed, recent theoretical work on unconstrained feature models has shown that gradient-based training exhibits strong implicit bias toward geometrically structured solutions, including simplex and orthogonal feature configurations, closely related to the neural collapse phenomenon observed in deep networks~\cite{e2022emergence,tirer2022extended,suken2023deep}. In particular, when multiple feature vectors are jointly optimized through a shared linear decoder, gradient descent tends to decorrelate these features and distribute them evenly in representation space. In our setting, each concept is represented by a shared latent sub-vector reused across all samples in which the concept appears, inducing a strong coupling between samples that share semantic structure. As a result, the implicit bias of the unconstrained feature model suggests that different concept subspaces are encouraged to become approximately decorrelated $\langle (y_{j,{k_1}})_{j \in \{1, ..., d\}}, (y_{j,{k_1}})_{j \in \{1, ..., d\}} \rangle \approx 0$ when $k_1 \neq k_2$, supporting stable semantic composition and reducing interference between concepts. This perspective offers a theoretical explanation for the fact that our framework encourages compositional generalization and aligns our framework with known neural collapse and geometry results in unconstrained feature models.

\paragraph{Feature-level modular editing.} Another potential application is editing via feature-level intervention, i.e.\ without modifying the prompt text. Indeed, if the concepts have been correctly disentangled, we can edit outputs by manipulating features in the sparse latent space: removing, swapping and inserting concepts could in principle be achieved by zeroing out or assigning learned values $y_{j,k}$ to the relevant sub-vectors. We have provided a detailed diagram of an editing workflow based on our decoder-only SSAE in Figure \ref{fig:money}. 

\paragraph{Encoders for SSAE.}\label{paragraph-encoders} Our SSAE framework also supports the use of an encoder. Indeed, we can view our sparse feature design as a procedure to construct a mask $\mathbf{M}$ which is a matrix of zeros and ones with the same dimensions as $\mathbf{Y}$, where coefficients are equal to 1 exactly at the indices where $\mathbf{Y}$ is trainable. We can then consider an encoder: $f_{\theta_1}: \mathbb{R}^{N} \mapsto \mathbb{R}^{d \cdot K}$ with any architecture, and train an encoder-decoder SSAE to reconstruct $\mathbf{X}$ by minimizing the loss: 
\begin{align*}
    \mathcal{L} = \mathbin\Vert \mathbf{X} - \mathbf{W}_2 [f_{\theta_1} (\mathbf{X}) \odot \mathbf{M}] \mathbin\Vert_2
\end{align*}
where $\odot$ is the element-wise matrix multiplication. The objective is also fully differentiable with respect to $\mathbf{W}_2$ and $\theta_1$. This offers additional capabilities compared to the decoder-only version. Indeed, to construct a new sparse latent representation $\mathbf{y}$, one does not need to start from an existing column of $\mathbf{Y}$, one can start from any $\mathbf{x}$ in the feature map or prompt embedding space, and apply the transformation $f_{\theta_1} (\mathbf{x}) \odot \mathbf{M}$ to arrive in the sparse latent space.

\begin{figure}[h!]
    \centering
    \begin{minipage}{0.30\textwidth}
        \centering
        \includegraphics[width=\textwidth]{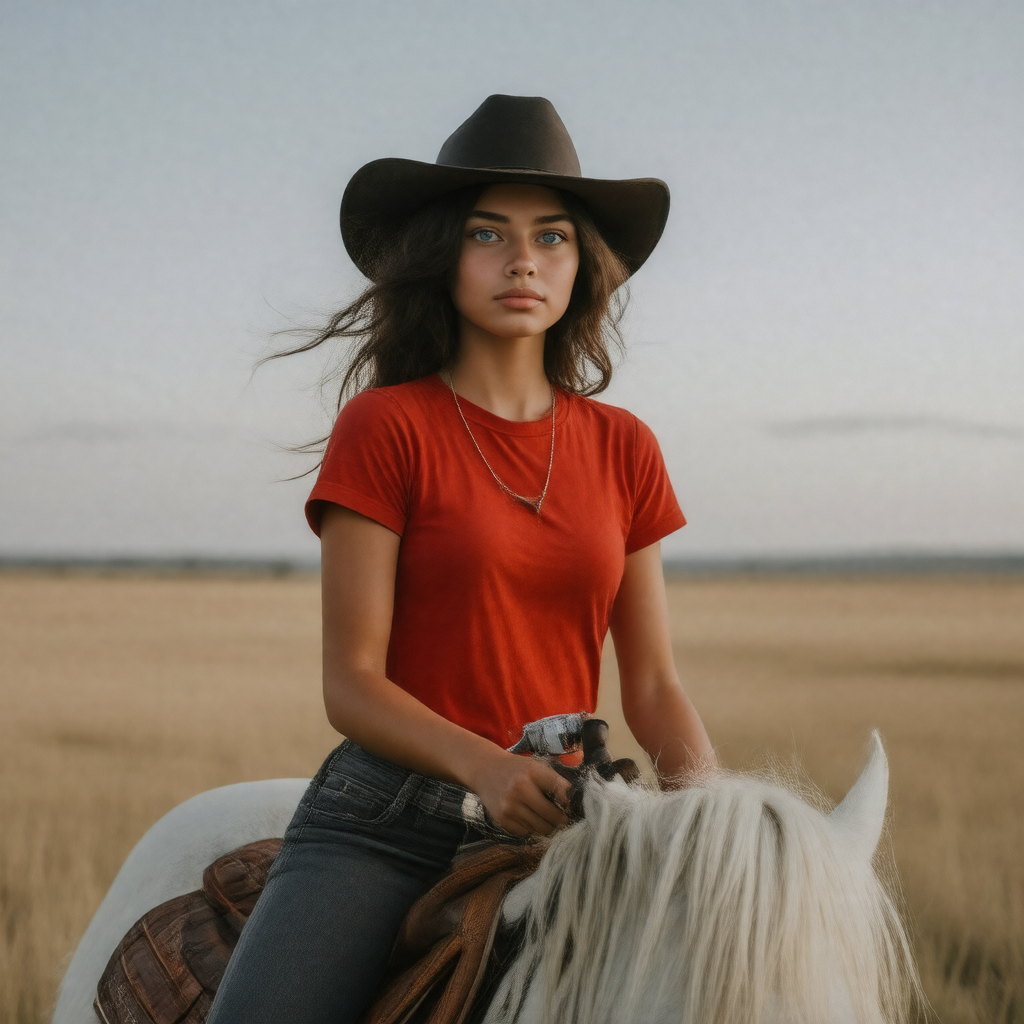}
        \label{fig:image5}
    \end{minipage}
    \hfill
    \begin{minipage}{0.30\textwidth}
        \centering
        \includegraphics[width=\textwidth]{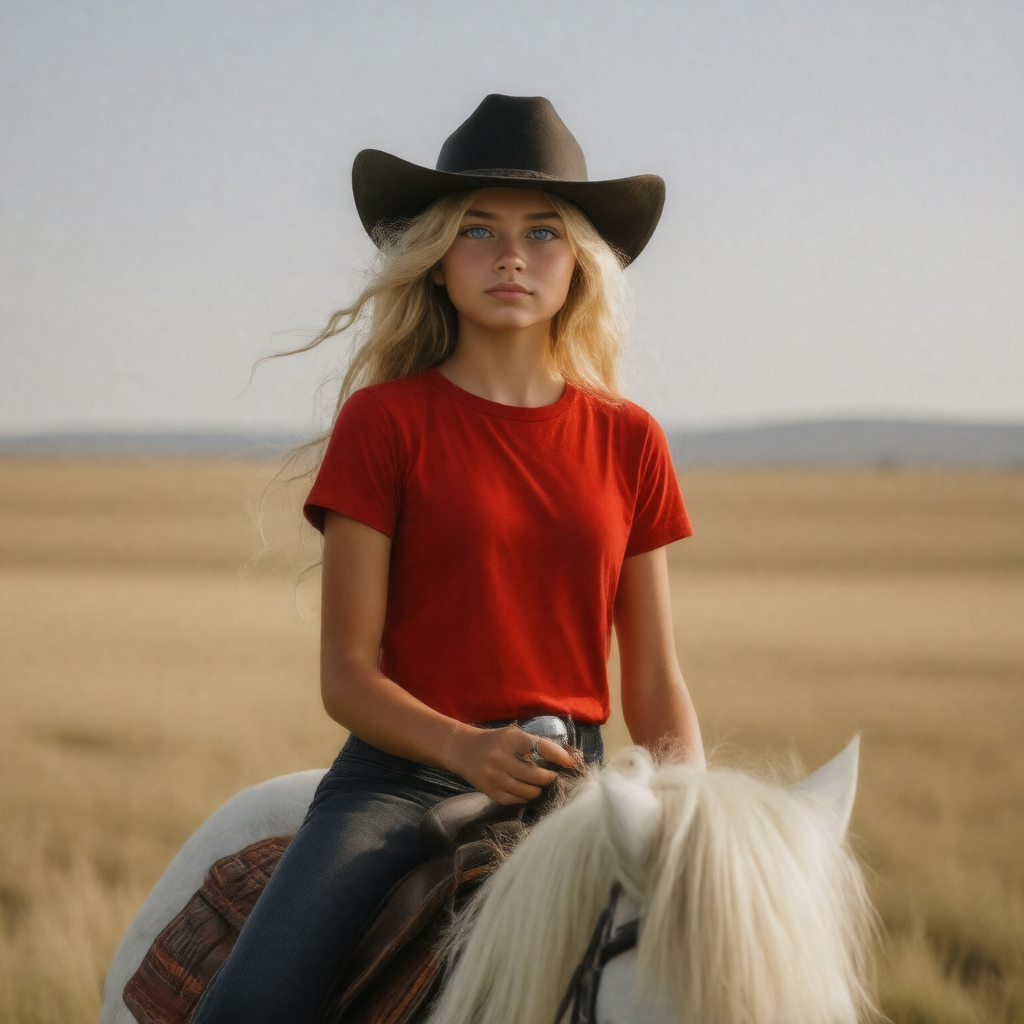}
        \label{fig:image6}
    \end{minipage}
    \caption{Initial prompt: "A brune girl with blue eyes on horseback across a plain, wearing a red t-shirt and a hat, holding a gun, looking in front of her."; then we perform the swap between "brune" and "blond" via our transformation in the sparse latent space learnt by our decoder-only SSAE.}
    \label{fig:brune_to_blond}
\end{figure}

\begin{figure}[h!]
    \centering
    \begin{minipage}{0.30\textwidth}
        \centering
        \includegraphics[width=\textwidth]{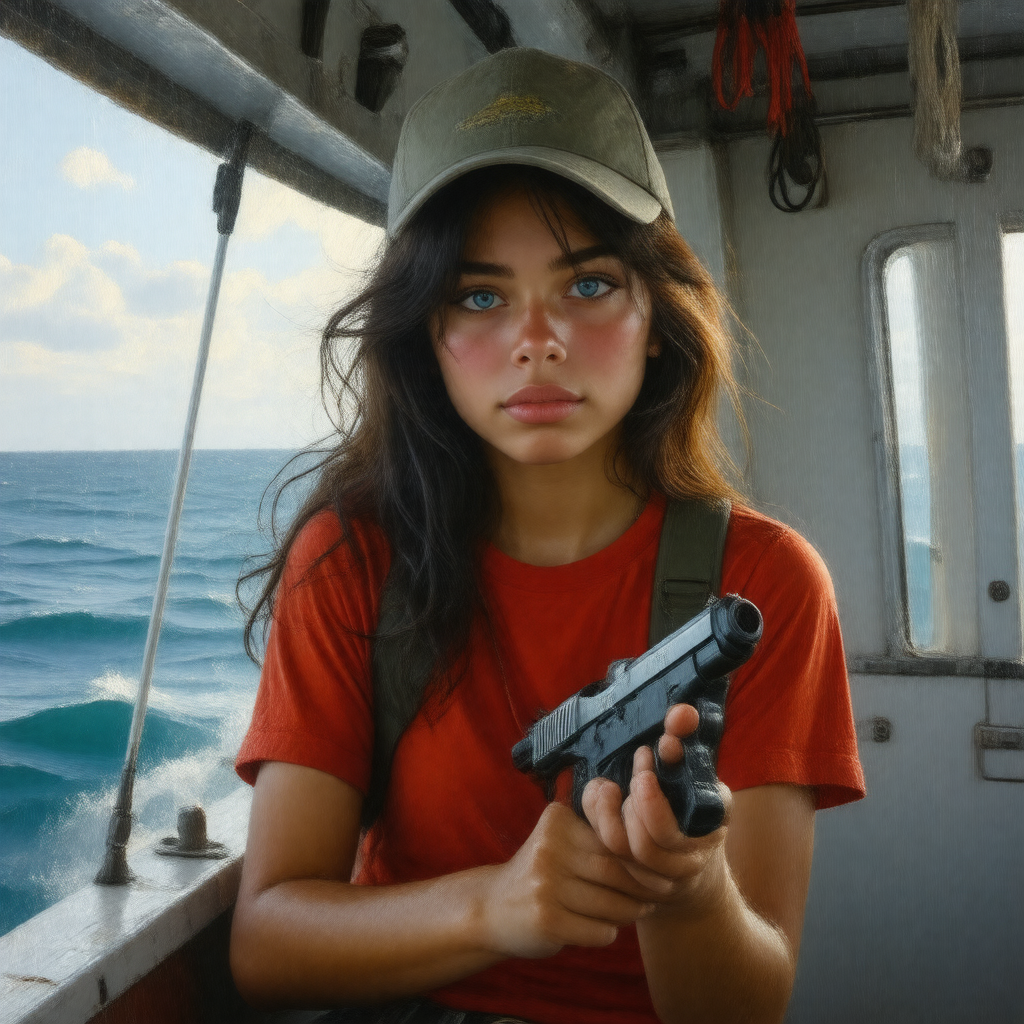}
        \label{fig:image7}
    \end{minipage}
    \hfill
    \begin{minipage}{0.30\textwidth}
        \centering
        \includegraphics[width=\textwidth]{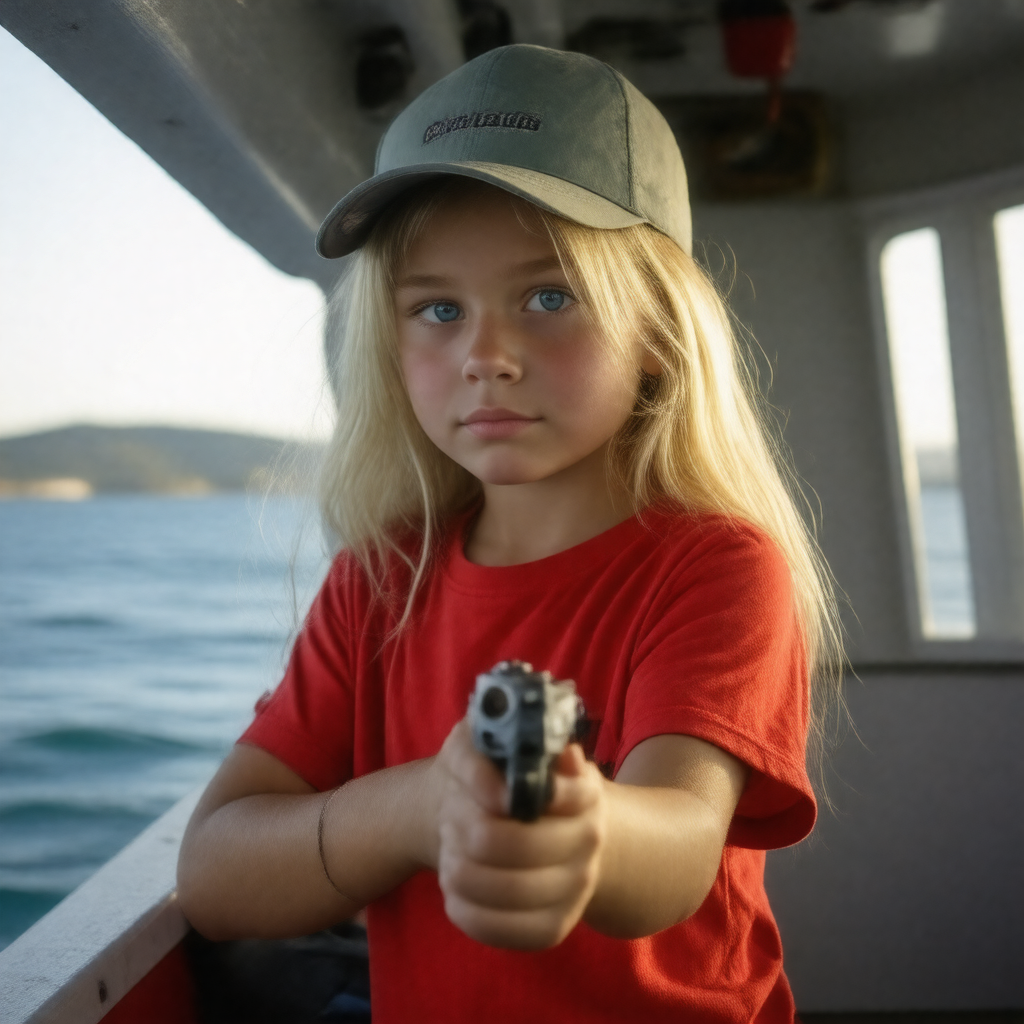}
        \label{fig:image8}
    \end{minipage}
    \caption{Initial prompt: "A brune girl with blue eyes on a boat, wearing a red t-shirt and a cap, holding a gun, looking in front of her."; then we perform the swap between "brune" and "blond" via our transformation in the sparse latent space learnt by our decoder-only SSAE.}
    \label{fig:brune_to_blond2}
\end{figure}

\section{Experiments.}

\paragraph{Compute and implementation details.}
All experiments were conducted on a single NVIDIA A10G GPU with 24~GB of memory. We used Stable Diffusion~3.5 Large Turbo~\cite{10.5555/3692070.3692573} for image generation, accessed through the HuggingFace \texttt{diffusers} library, and employed it strictly for inference without fine-tuning or modifying the diffusion model parameters. Stable Diffusion~3.5 uses a T5-based text encoder for prompt conditioning, for which prompt embeddings have dimensionality $\sim$1.3M; all prompt embeddings used in our experiments were extracted from this frozen encoder~\cite{raffel2020t5}. The diffusion model was loaded using 4-bit quantization with NF4 weights via \texttt{bitsandbytes}, using \texttt{bfloat16} for computation.

The supervised sparse auto-encoder was trained in the decoder-only configuration (linear decoder and sparse concept matrix) on approximately 1500 prompt embeddings, and we set the concept subspace dimension to $d = 10$. Training completed in approximately 12 minutes on a single GPU. All image edits were performed at inference time by directly modifying prompt embeddings, with no gradient-based optimization during generation. Experiments were run using PyTorch with CUDA~12.4.

The code to reproduce the experiments can be found at \url{https://github.com/ouns972/decoder-only-ssae}.

\paragraph{Training set.} We constructed a concept dictionary of visual attributes, including physical features (\texttt{blond hair}, \texttt{brune hair}, \texttt{blue eyes}, \texttt{black eyes}), poses (\texttt{seated}, \texttt{on horseback}, \texttt{standing}), object presence (\texttt{gun}, \texttt{coffee cup}, \texttt{coca-cola}), and environment (\texttt{car}, \texttt{boat}, \texttt{bar}, \texttt{street}). We generated a dataset of 1500 prompts with varied concept combinations and extracted the corresponding Stable Diffusion 3.5 prompt embeddings via the T5 text encoder. To showcase compositional generalization, we specifically designed a training dataset where some concepts are never seen together.

%, for example with prompts where "blond hair, blue eyes" characters are depicted in situations in which "brune hair, blue eyes" characters are never seen. 

\paragraph{Sparse feature design.} We applied our sparse feature design procedure by assigning to each concept a dedicated block in the sparse latent space: concept sub-vectors are trainable only for prompts containing that concept, and zero otherwise. We then trained a decoder-only SSAE to reconstruct the prompt embeddings. 

\paragraph{Compositional generalisation.} After training, we performed three types of operations on the trained sparse feature matrix:
\begin{itemize}
    \item Swap between concepts $c_{k_1} \to c_{k_2}$: setting to zero all components with indices $(j, k_1)$  and setting all components with indices $(j, k_2)$ to learned values $y_{j, k_2}$ for $j \in \{1, ..., d\}$;
    \item Removal of concept $c_k$: setting to zero all components with indices $(j, k)$ for $j \in \{1, ..., d\}$;
    \item Insertion of concept $c_{k'}$ setting all components with indices $(j, k')$ to learned values $y_{j, k'}$ for $j \in \{1, ..., d\}$.
\end{itemize}

\begin{figure*}[h]
    \centering
    \begin{minipage}{0.30\textwidth}
        \centering
        \includegraphics[width=\textwidth]{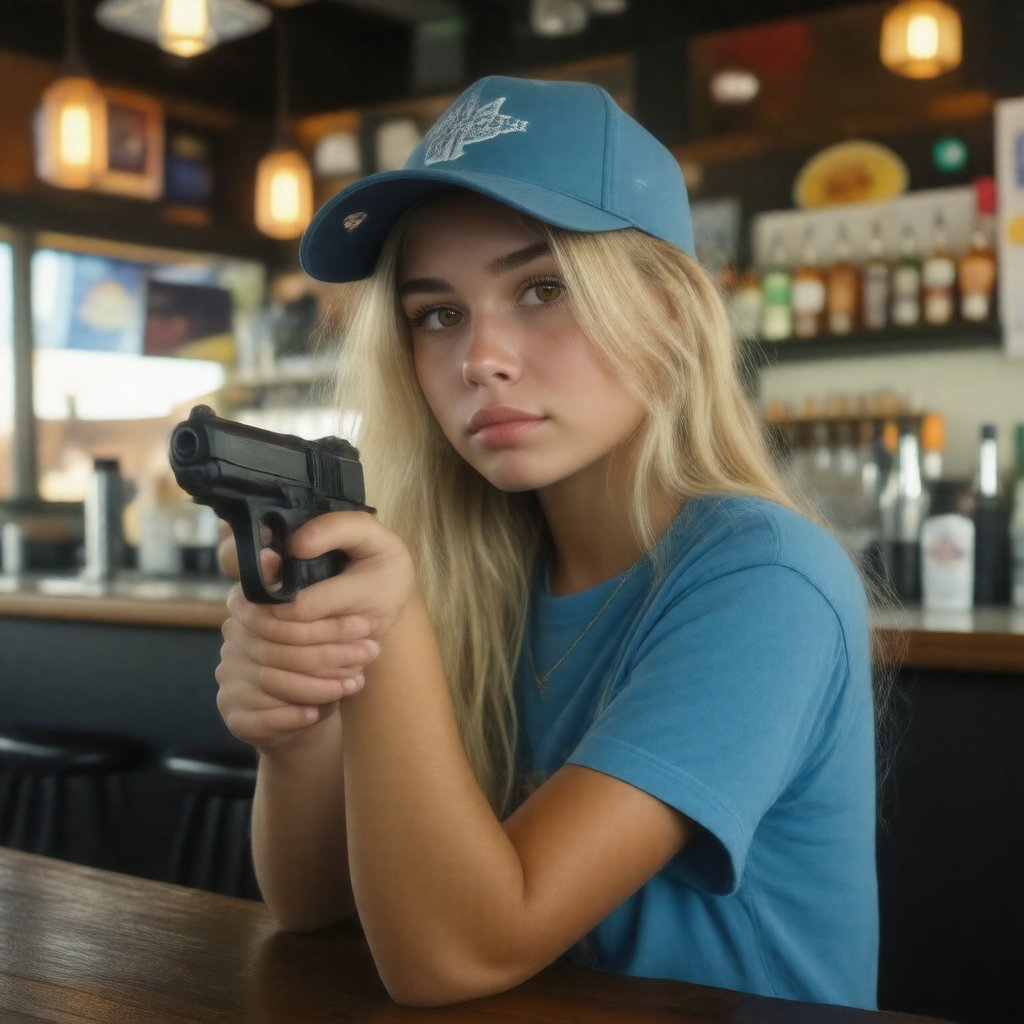}
        \label{fig:image7}
    \end{minipage}
    \begin{minipage}{0.30\textwidth}
        \centering
        \includegraphics[width=\textwidth]{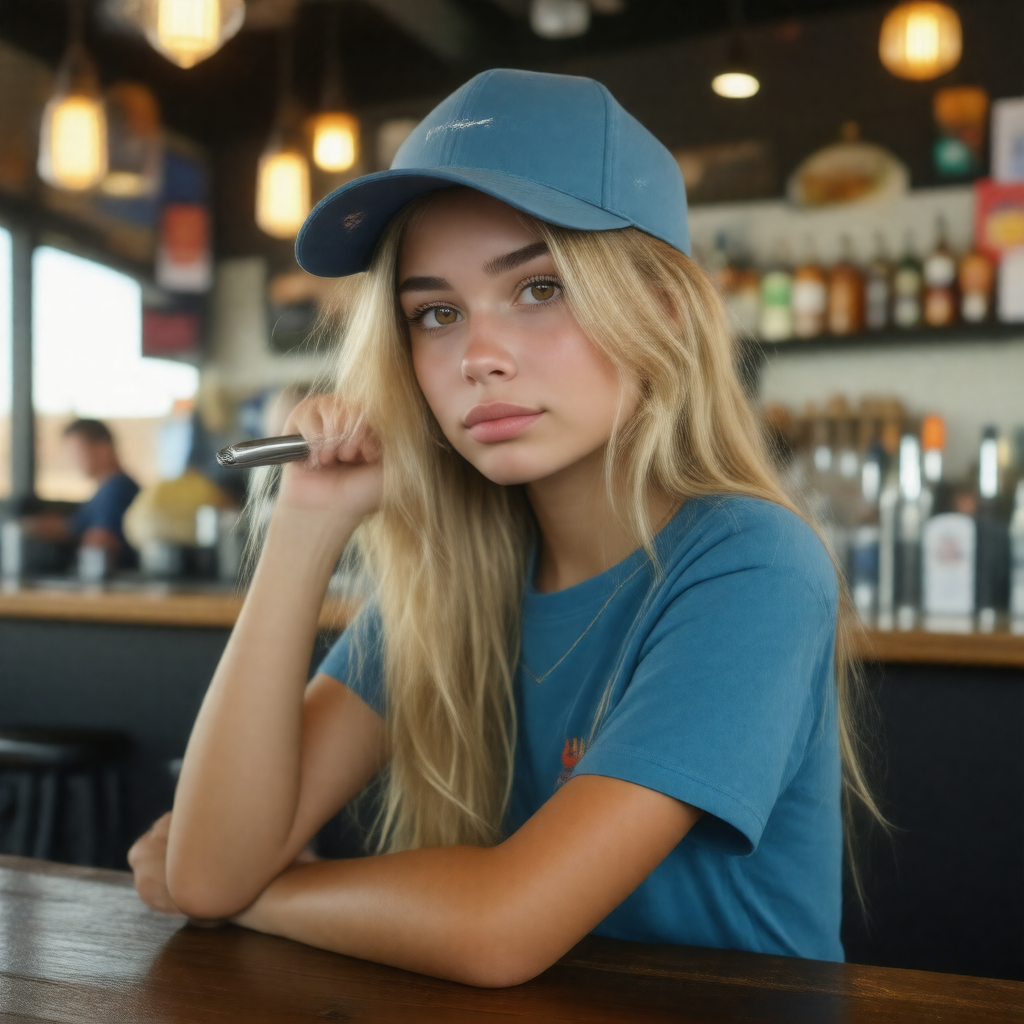}
        \label{fig:image7}
    \end{minipage}
    \\
    \begin{minipage}{0.30\textwidth}
        \centering
        \includegraphics[width=\textwidth]{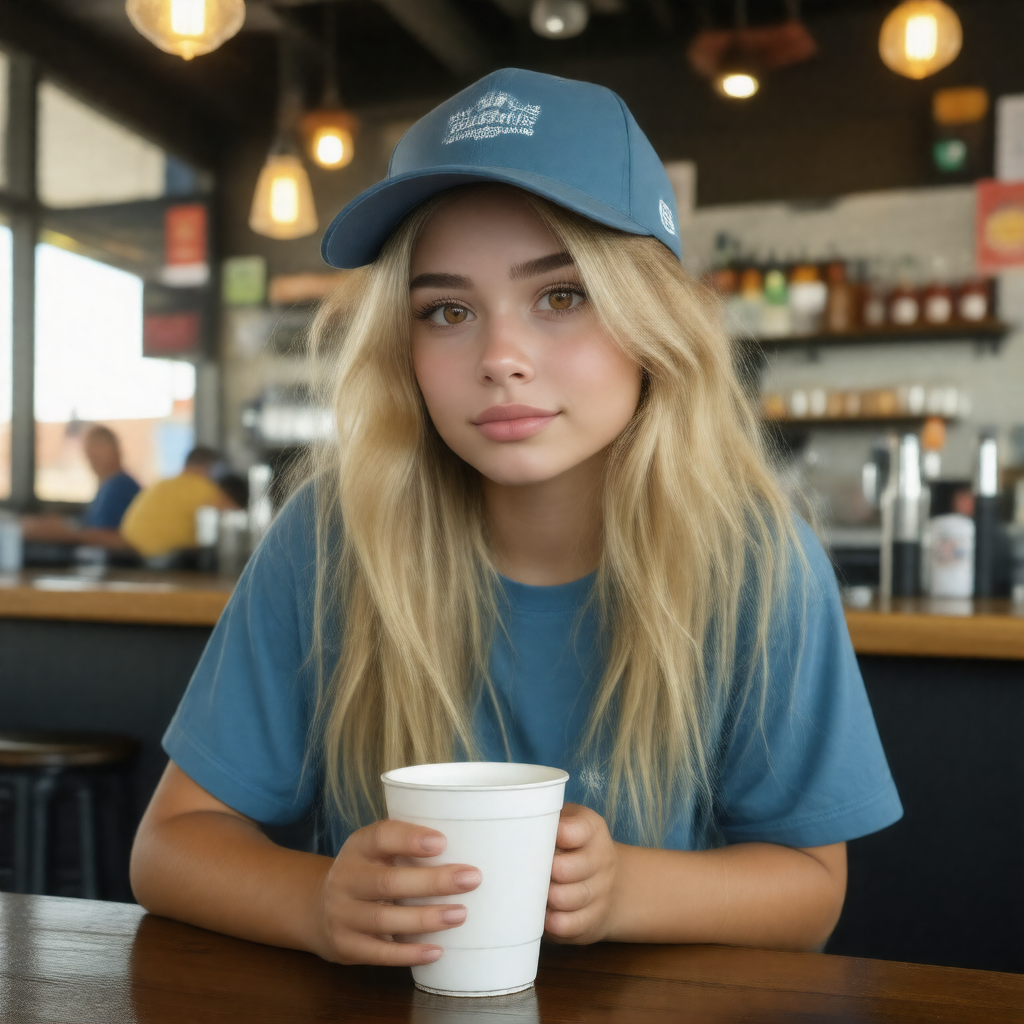}
        \label{fig:image8}
    \end{minipage}
    \begin{minipage}{0.30\textwidth}
        \centering
        \includegraphics[width=\textwidth]{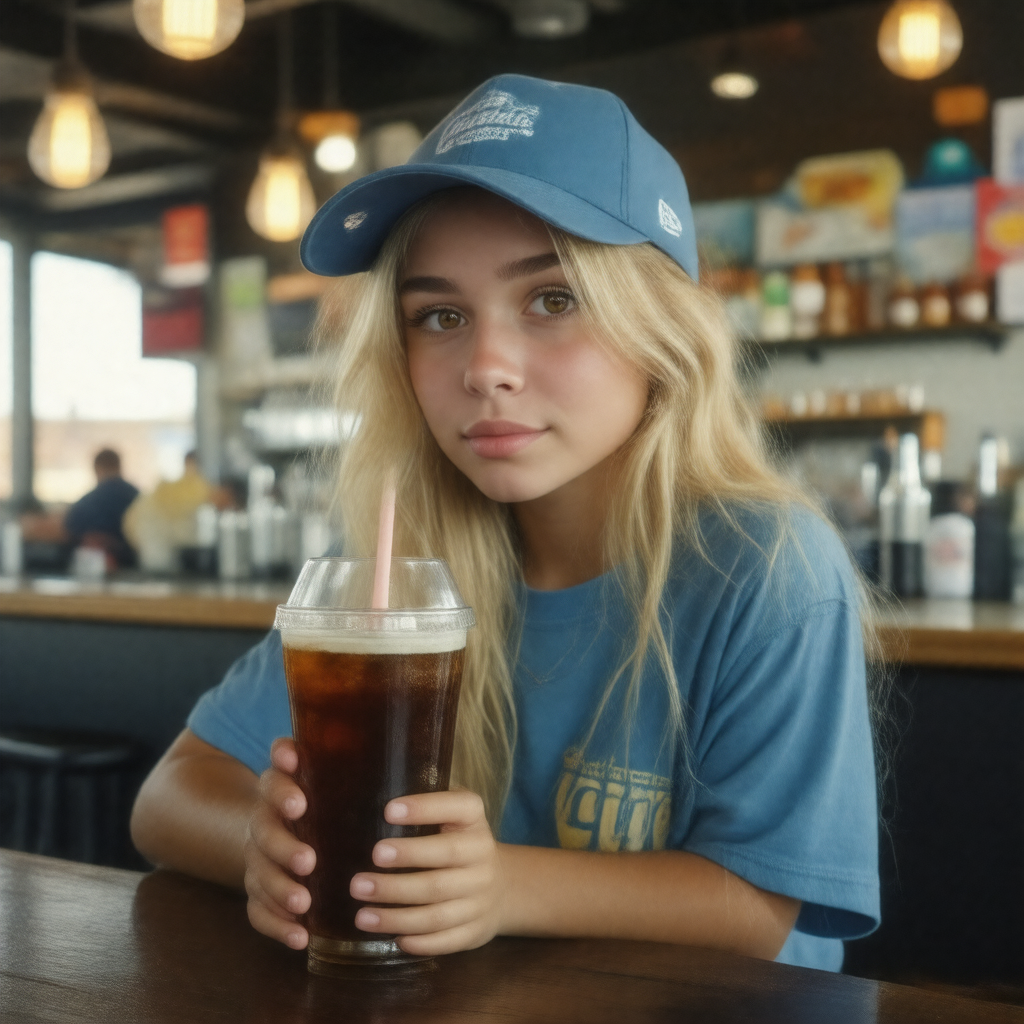}
        \label{fig:image8}
    \end{minipage}
    \caption{Initial prompt: "A blong girl with brown eyes sitting at a bar, wearing a blue t-shirt and a baseball cap, holding a gun, looking in front of her." (top-left), then we applied our transformation in the sparse latent space learnt by our decoder-only SSAE to successively: remove the concept of "holding a gun" (top-right), insert the concept of "holding a coffee" (bottom-left), swap it with "holding a coca-cola" (bottom-right), showcasing compositional generalisation.}
    \label{fig:bar}
\end{figure*}

\begin{figure*}[t]
    \centering
    \begin{minipage}{0.30\textwidth}
        \centering
        \includegraphics[width=\textwidth]{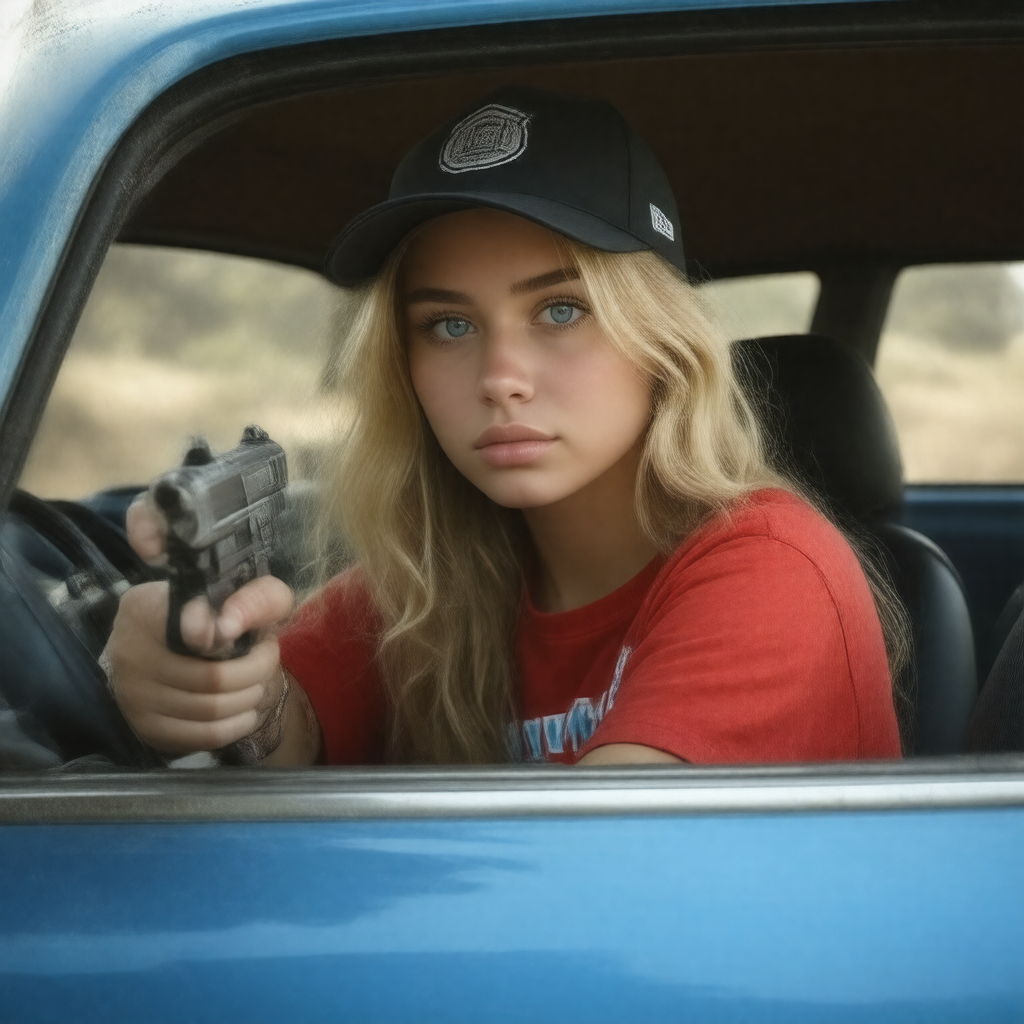}
        \label{fig:image7}
    \end{minipage}
    \begin{minipage}{0.30\textwidth}
        \centering
        \includegraphics[width=\textwidth]{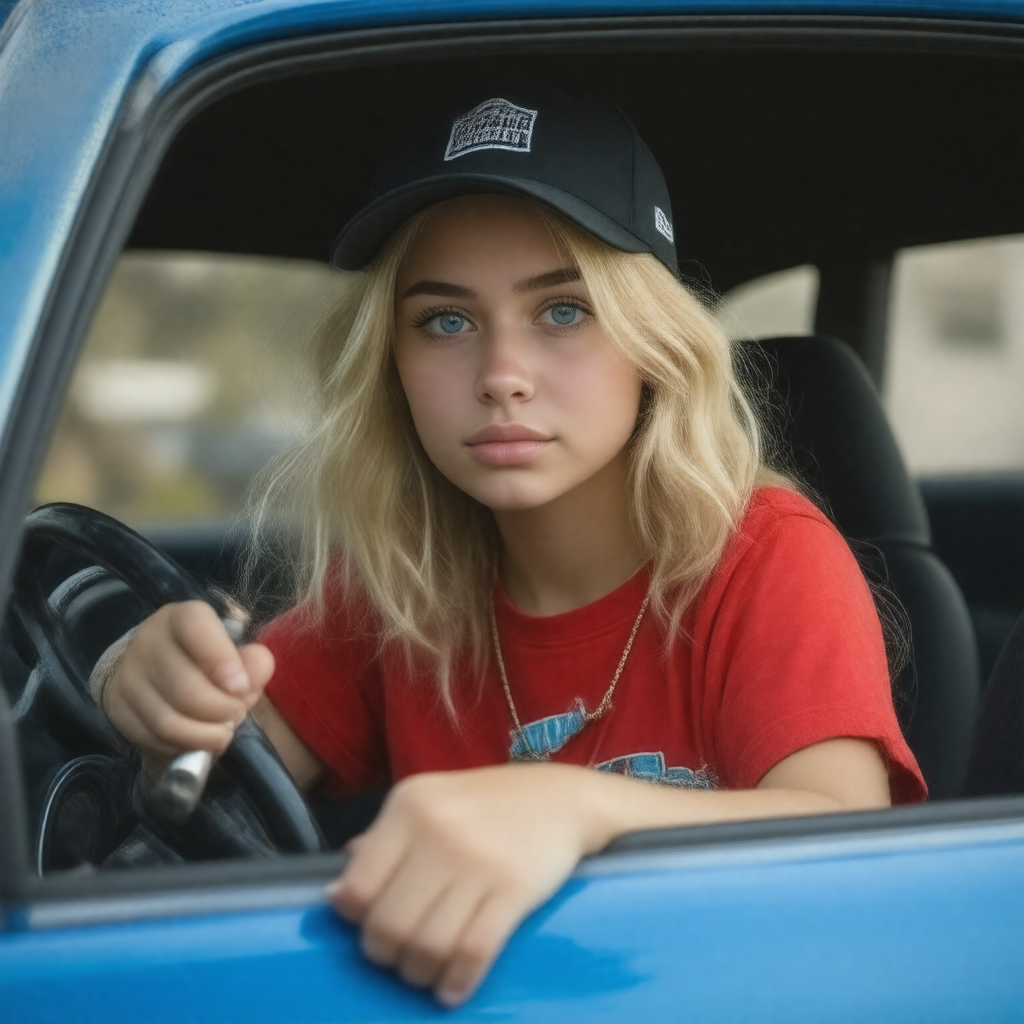}
        \label{fig:image8}
    \end{minipage}
    \begin{minipage}{0.30\textwidth}
        \centering
        \includegraphics[width=\textwidth]{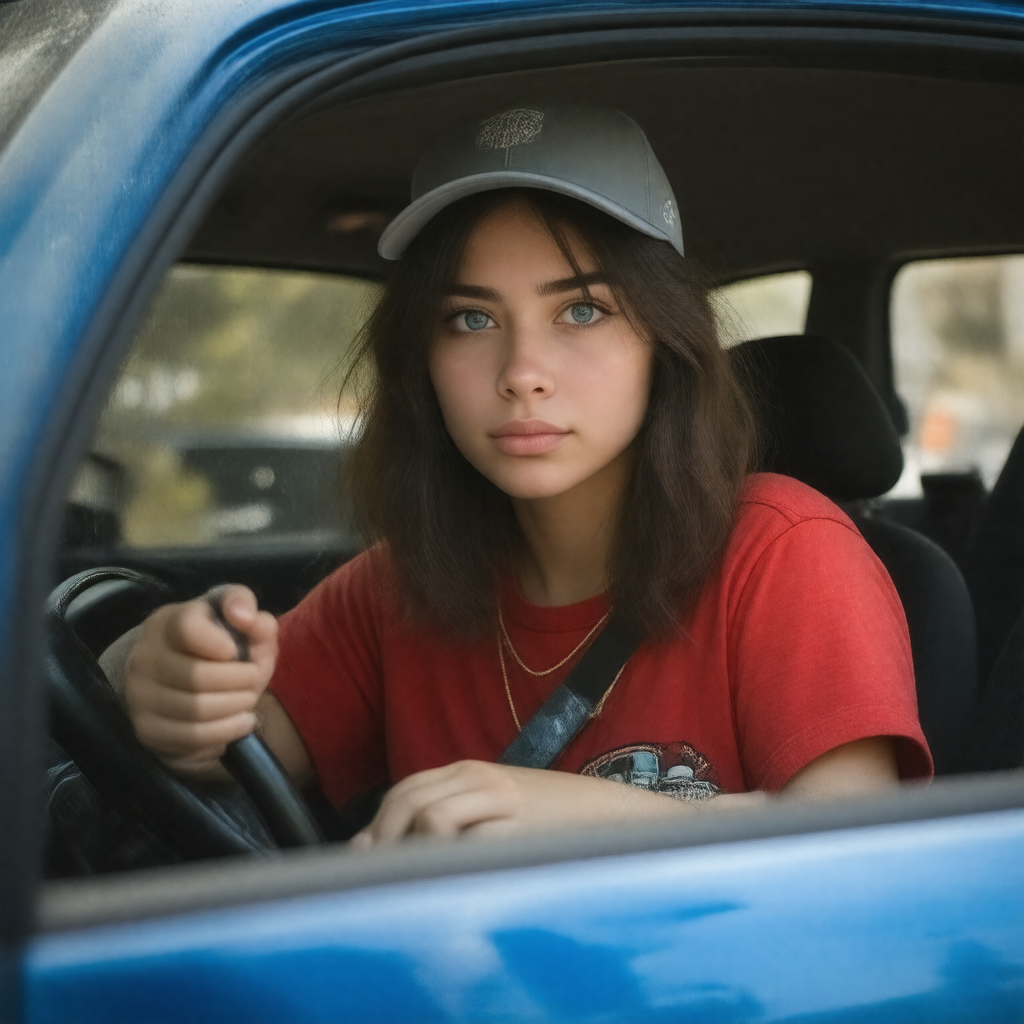}
        \label{fig:image8}
    \end{minipage}
    \caption{Initial prompt: "A blond girl with blue eyes in a car, wearing a red t-shirt and a cap, holding a gun, looking in front of her." (left), then we applied our transformation in the sparse latent space learnt by our decoder-only SSAE to successively: remove the concept of "holding a gun" (middle), swap the concepts of "blond hair" and "brune hair"(right), showcasing compositional generalisation.}
    \label{fig:car}
\end{figure*}

We performed semantic composition by starting from columns in $\mathbf{Y}$ and applying such transformations successively, as illustrated Figures \ref{fig:bar} and \ref{fig:car}. This procedure thus produced new sparse latent representations $\mathbf{y}$ that did not appear in our training set.  By computing $\mathbf{W_2} \sigma (\mathbf{y})$, we obtained new reconstructed prompt embeddings $\hat{\mathbf{x}}$, that corresponds to prompts never written in the training set.  We then used the prompt embeddings to perform inferences with Stable Diffusion 3.5, and inspect the output images.

For example, as illustrated Figures \ref{fig:brune_to_blond} and \ref{fig:brune_to_blond2}, we picked prompts with "brune hair, blue eyes" characters, in configurations for which our training set did not contain any "blond hair, blue eyes" characters. Inspecting the output images, we observed, as expected, characters depicted in the same original configuration but with blond hair and blue eyes. This qualitative evaluation showcases that the decoder-only SSAE has correctly isolated the concept of "brune hair" and "blond hair" since a character with blond hair and blue eyes was never seen in these situations in the training data. To stress-test the statistical significance of these results, on this specific task of changing hair color, we visually inspected $>$50 images and found that it correctly changes the hair color 100\% of the time. We note however that this may be an "easy" task, i.e.\ where the concepts are easily linearly accessible in the prompt embedding space. Additional experimental results can be found in the Appendix.

Another limitation worth mentioning is that the transformations are not always neutral w.r.t.\ the rest of the image (they do not only visually modify the targeted concept), but the coarse-grained behavior showcases that our method demonstrate compositional generalisation in terms of (high-level) semantics. We argue that the success of our method on this task with only 1500 prompts, and a low concept subspace dimension $d =10$ suggests that our method has huge potential in terms of scalability. We didn't push the experimental exploration to its full potential due to computational constraints but we indeed think that an avenue to obtain fine-grained disentanglement of concepts would be to increase the concept subspace dimension $d$, and the number of prompts $n$ in which the concepts are seen in different combinations. As mentioned above, training can be batched with respect to the number of prompts and gradient updates for a decoder-only SSAE scale linearly with $d$, which offers a lot of potential.

\paragraph{Feature-level modular editing.} Although we do not claim that the current experimental results are on par with state-of-the-art editing methods for image diffusion models, we argue that our method yields a feature-level modular editing workflow (see Figure \ref{fig:money}). Scaling up the experiments, i.e. training on many more prompts and increasing the dimensionality of each concept sub-vector is a clear avenue for benchmarking this new method against existing editing workflows. Our preliminary experimental results do indicate that concept-aligned sparse vectors can reliably induce semantically targeted edits. The model generalizes to unseen combinations, supporting compositional modular edits.

\section{Related Works}

\paragraph{Mechanistic interpretability.}
Mechanistic interpretability research has increasingly turned to sparse autoencoders (SAEs) to extract human-understandable features from complex models. In large language models (LLMs), unsupervised SAEs have been used to address superposition by identifying latent directions that are more \textit{monosemantic} than raw neurons or PCA components~\cite{huben2024sparse}. These features often correspond to meaningful behaviors and enable fine-grained causal analysis. Indeed, for Transformer activations in large language models, unsupervised SAEs trained on residual stream or attention outputs help improve interpretability and enabling controllable interventions (\cite{bricken2023towards, kissane2024interpreting, makelov2024towards, gao2025scaling}).  In diffusion models, training SAEs on U-Net activations has yielded latent units associated with high-level generative factors such as image layout, lighting, and object presence, and has found that spatially localized, semantic features—such as color, composition, and detail—that can be causally intervened upon \citet{surkov2025one}. On prompt embeddings, \citet{tian2025sparse} used SAEs on CLIP embeddings to identify semantically coherent latent components, enabling precise prompt editing and conditional generation control. Critically, interventions along these sparse directions produce predictable changes in the model output, revealing their functional role~\cite{bhalla2024interpreting}. As a result, SAEs are increasingly used to move from \textit{post hoc probing} to \textit{active steering} of model behavior. 

Tangentially, linear probing is a supervised method that is widely-used for evaluating the information encoded in learned representations. It measures whether a target attribute (e.g., object class, sentiment, or part-of-speech tag) can be predicted from a frozen embedding using a simple linear classifier~\cite{alain2016understanding,hewitt2019structural}. High probe accuracy indicates that the attribute is linearly accessible, suggesting it is explicitly represented in the embedding. However, both unsupervised SAEs and linear probes come with important limitations. Linear probes can detect whether a concept is encoded, but do not isolate it causally. SAEs, while more powerful, may still entangle multiple concepts in a single neuron or split one concept across several~\cite{chanin2026a}. Additionally, reconstruction loss may bias SAEs toward frequent patterns, missing semantically rare but meaningful features~\cite{gao2025scaling}. Finally, large SAEs often yield unstable decompositions with no canonical alignment between neurons and concepts~\cite{leask2025sparse}.

Our SSAE framework sits at the intersection of SAEs and linear probing, borrowing supervision from linear probing and sparsity from  SAEs.  SSAEs aim to resolve the challenges with both techniques by incorporating concept-level guidance. Instead of discovering structure post hoc, SSAEs define interpretable features \textit{a priori}, assigning each concept to its own sparse subspace. Recent and concurrent works like SAEmnesia~\cite{cassano2025saemnesia}, AlignSAE~\cite{yang2025alignsae} and CASL \cite{he2026casl} also show that the supervised approach improves alignment, reduces concept splitting, and enables modular editing. For instance, SAEmnesia was able to erase specific features from a diffusion model by zeroing a single unit, while AlignSAE fine-tuned sparse features to match human-interpretable ontologies and CASL combined unsupervised SAEs and supervised concept alignment to enable modular editing. These methods, similarly to ours, allow edits to be compositional and localized in latent space---a capability that traditional SAEs and probes lack. Our work builds on this direction, proposing a decoder-only SSAE based in the theory of  unconstrained feature models (an angle that none of these works take), opening the way to scalable, structured, and semantically grounded intervention in large models.

\paragraph{Dictionary learning and block-sparse representations.}
Dictionary learning aims to represent data as sparse linear combinations of learned atoms, typically by optimizing a reconstruction objective with sparsity-inducing regularization such as an $\ell_1$ penalty or greedy pursuit algorithms~\cite{elad2010sparse,mairal2014sparse}. Extensions to structured and block-sparse settings introduce group-level sparsity patterns to capture correlations among features~\cite{jenatton2011structured}. In these approaches, sparse codes are generally inferred independently for each data point, and the learned structure reflects statistical regularities in the data rather than explicit semantic supervision. By contrast, the SSAE framework studied in this work defines the sparsity pattern a priori at the level of human-interpretable concepts and learns shared latent sub-vectors across all samples in which a concept appears, enabling explicit semantic alignment and compositional feature-level interventions.

\paragraph{Editing in text-to-image diffusion models.}
Text-to-image diffusion models support a range of editing techniques operating at different levels of the generation pipeline. \emph{Prompt-to-prompt editing} modifies cross-attention layers during generation to steer outputs based on textual edits~\cite{Hertz2023}. \emph{Textual inversion} learns pseudo-word embeddings to capture new visual concepts from few-shot image examples~\cite{Gal2023}, while \emph{concept erasure} techniques like ESD fine-tune model weights to eliminate specific content~\cite{Gandikota2023}. Latent-based strategies such as \emph{Imagic} optimize prompt embeddings to better align with target edits while preserving image identity~\cite{Kawar2023}. In contrast to these methods, our approach edits the \emph{prompt embedding space} directly via decoded sparse features. By aligning sparse latent blocks with semantic concepts, our model supports compositional interventions (e.g., removing ``gun'' and adding ``blond hair'') without modifying the model or relying on gradient-based optimization at inference. Unlike prompt-to-prompt and textual inversion, which treat prompt embeddings as atomic vectors, our method introduces an interpretable, structured basis over the prompt space, enabling modular edits grounded in human-understandable features.

%\begin{figure}[h!]
 %   \centering
  %  \begin{minipage}{0.45\textwidth}
   %     \centering
    %    \includegraphics[width=\textwidth]{image_blond_ouns.png}
     %   \label{fig:image1}
    %\end{minipage}
    %\hfill
    %\begin{minipage}{0.45\textwidth}
     %   \centering
      %  \includegraphics[width=\textwidth]{image_blond_switched_to_brune_reconstructed_by_hand_ouns.png}
      %  \label{fig:image2}
    %\end{minipage}
    %\caption{Initial prompt: "A blond girl with blue eyes sitting at a coffee shop, wearing a red t-shirt and a cap, holding a coffee, looking in front of her.", then we set the components corresponding to the feature "blond girl" to zero and we activate the feature "brune girl" by setting the corresponding components to the values learnt by the supervised sparse auto-encoder.}
%\end{figure}

%\begin{figure}[h!]
 %   \centering
  %  \begin{minipage}{0.45\textwidth}
   %     \centering
    %    \includegraphics[width=\textwidth]{image_blond_ouns2.png}
     %   \label{fig:image3}
    %\end{minipage}
    %\hfill
    %\begin{minipage}{0.45\textwidth}
    %    \centering
     %   \includegraphics[width=\textwidth]{image_blond_switched_to_brune_reconstructed_by_hand_ouns2.png}
      %  \label{fig:image4}
    %\end{minipage}
    %\caption{Initial prompt: "A blond girl with blue eyes at a market, wearing a black t-shirt and a cap, holding a gun, looking to the left.", then we set the components corresponding to the feature "blond girl" to zero and we activate the feature "brune girl" by setting the corresponding components to the values learnt by the supervised sparse auto-encoder.}
%\end{figure}

\section{Conclusion and Limitations}

We presented a supervised sparse auto-encoder framework for aligning interpretable concept vectors with internal representations, applicable to foundational models, and we verified the soundness of our methodology on prompt embeddings for Stable Diffusion 3.5. By directly supervising a sparse latent space with human-defined concept structure, and training a decoder-only architecture, our method avoids many of the challenges of unsupervised SAEs: it requires no L1 regularization, no encoder, and yields concept-aligned directions by design. In its simplest form, our methodology readily supports compositional generalisation which provides a new lens into interpretability. The methodology can be further enriched by incorporating an encoder, which allows to inspect any instance at test time and intervene at feature-level by manipulating the sparse latent space. 

Our preliminary experimental results showcase the potential of this approach to enable intrepretability through semantic composition and feature-level modular editing. This opens the door to efficient and interpretable editing pipelines grounded in feature-level control.

\subsection*{Limitations and Future Work}
While promising, our method also comes with limitations:
\begin{itemize}
    \item \textbf{Predefined concept scope.} The model can only edit concepts that were included in the supervised dictionary. It does not support discovery of new, emergent features.
    \item \textbf{Manual concept encoding.} Building high-quality concept dictionaries requires human input and may not scale easily to thousands of attributes. We note however that this could be automated by large (vision) language models.
    \item \textbf{Limited experimental set-up.} Due to computational constraints, we have not performed large-scale experiments that would enable to reach the full potential of the methodology, especially in terms of editing, and compare againsts state-of-the-art methods. Our experimental results in their current state are merely evidence of soundness of the methodology. For the same reason, we have not trained an encoder for our SSAE: this would be beneficial to assess the performance of the method on inputs unrelated to any training examples.
    \item \textbf{Limited evaluation domain.} We only tested the method on prompt embeddings from Stable Diffusion 3.5. Future work should apply the same architecture to U-Net feature maps or transformer hidden states to assess generality, across application domains (text generation, image generation, video generation etc).
\end{itemize}

Despite these limitations, our method offers a scalable approach for interpretable control in large foundation models. We believe this structured, concept-supervised approach can serve as a useful alternative to unsupervised SAEs in applications where feature-level modularity, editability, and semantic alignment are desired.

%\section{Impact Statement}
%This paper presents work whose goal is to advance the field of machine learning. There are many potential societal consequences of our work, none of which we feel must be specifically highlighted here.

\bibliography{example_paper}

@book{elad2010sparse,
  title={Sparse and Redundant Representations: From Theory to Applications in Signal and Image Processing},
  author={Elad, Michael},
  year={2010},
  publisher={Springer}
}

@article{mairal2014sparse,
  title={Sparse modeling for image and vision processing},
  author={Mairal, Julien and Bach, Francis and Ponce, Jean},
  journal={Foundations and Trends in Computer Graphics and Vision},
  volume={8},
  number={2--3},
  pages={85--283},
  year={2014},
  publisher={Now Publishers}
}

@article{jenatton2011structured,
  title={Structured sparse principal component analysis},
  author={Jenatton, Rodolphe and Obozinski, Guillaume and Bach, Francis},
  journal={Journal of Machine Learning Research},
  volume={12},
  pages={277--312},
  year={2011}
}

@article{e2022emergence,
  title={On the emergence of simplex symmetry in the final and penultimate layers of neural network classifiers},
  author={E, Weinan and Wojtowytsch, Stephan},
  journal={Mathematical and Scientific Machine Learning},
  pages={270--290},
  year={2022}
}

@inproceedings{tirer2022extended,
  title={Extended unconstrained features model and the neural collapse phenomenon},
  author={Tirer, Tsvi and Bruna, Joan},
  booktitle={Advances in Neural Information Processing Systems},
  year={2022}
}

@misc{he2026casl,
  title        = {CASL: Concept-Aligned Sparse Latents for Interpreting Diffusion Models},
  author       = {Zhenghao He and Guangzhi Xiong and Boyang Wang and Sanchit Sinha and Aidong Zhang},
  year         = {2026},
  howpublished = {arXiv preprint arXiv:2601.15441},
}

@inproceedings{
gao2025scaling,
title={Scaling and evaluating sparse autoencoders},
author={Leo Gao and Tom Dupre la Tour and Henk Tillman and Gabriel Goh and Rajan Troll and Alec Radford and Ilya Sutskever and Jan Leike and Jeffrey Wu},
booktitle={The Thirteenth International Conference on Learning Representations},
year={2025},
url={https://openreview.net/forum?id=tcsZt9ZNKD}
}

@inproceedings{
makelov2024towards,
title={Towards Principled Evaluations of Sparse Autoencoders for Interpretability and Control},
author={Aleksandar Makelov and Georg Lange and Neel Nanda},
booktitle={ICLR 2024 Workshop on Secure and Trustworthy Large Language Models},
year={2024},
url={https://openreview.net/forum?id=MHIX9H8aYF}
}

@article{bricken2023towards,
  title={Towards monosemanticity: Decomposing language models with dictionary learning},
  author={Bricken, Trenton and Templeton, Adly and Batson, Joshua and Chen, Brian and Jermyn, Adam and Conerly, Tom and Turner, Nick and Anil, Cem and Denison, Carson and Askell, Amanda and others},
  journal={Transformer Circuits Thread},
  volume={2},
  year={2023}
}

@inproceedings{kissane2024interpreting,
  title={Interpreting attention layer outputs with sparse autoencoders},
  author={Kissane, Connor and Krzyzanowski, Robert and Bloom, Joseph Isaac and Conmy, Arthur and Nanda, Neel},
  booktitle={Mechanistic Interpretability Workshop at ICLR},
  year={2024}
}

@article{surkov2025one,
  title={One-step is enough: Sparse autoencoders for text-to-image diffusion models},
  author={Surkov, Viacheslav and Wendler, Chris and Mari, Antonio and Terekhov, Mikhail and Deschenaux, Justin and West, Robert and Gulcehre, Caglar and Bau, David},
  journal={arXiv preprint arXiv:2410.22366},
  year={2025}
}

@article{tian2025sparse,
  title={Sparse autoencoder as a zero-shot classifier for concept erasing in text-to-image diffusion models},
  author={Tian, Zhihua and Nan, Sirun and Xu, Ming and Zhai, Shengfang and Qu, Wenjie and Liu, Jian and Jia, Ruoxi and Zhang, Jiaheng},
  journal={arXiv preprint arXiv:2503.09446},
  year={2025}
}

@article{smith2025negative,
  title={Negative results for sparse autoencoders on downstream tasks and deprioritising sae research (mechanistic interpretability team progress update)},
  author={Smith, Lewis and Rajamanoharan, Sen and Conmy, Arthur and McDougall, Callum and Kramar, Janos and Lieberum, Tom and Shah, Rohin and Nanda, Neel},
  journal={DeepMind Safety Research},
  year={2025},
  note={\url{https://deepmindsafetyresearch.medium.com/negative-results-for-sparse-autoencoders-on-downstream-tasks-and-deprioritising-sae-researc}}
}

@inproceedings{
huben2024sparse,
title={Sparse Autoencoders Find Highly Interpretable Features in Language Models},
author={Robert Huben and Hoagy Cunningham and Logan Riggs Smith and Aidan Ewart and Lee Sharkey},
booktitle={The Twelfth International Conference on Learning Representations},
year={2024},
url={https://openreview.net/forum?id=F76bwRSLeK}
}

@inproceedings{
bhalla2024interpreting,
title={Interpreting {CLIP} with Sparse Linear Concept Embeddings (SpLi{CE})},
author={Usha Bhalla and Alex Oesterling and Suraj Srinivas and Flavio Calmon and Himabindu Lakkaraju},
booktitle={The Thirty-eighth Annual Conference on Neural Information Processing Systems},
year={2024},
url={https://openreview.net/forum?id=7UyBKTFrtd}
}

@inproceedings{
chanin2026a,
title={A is for Absorption: Studying Feature Splitting and Absorption in Sparse Autoencoders},
author={David Chanin and James Wilken-Smith and Tom{\'a}{\v{s}} Dulka and Hardik Bhatnagar and Satvik Golechha and Joseph Isaac Bloom},
booktitle={The Thirty-ninth Annual Conference on Neural Information Processing Systems},
year={2026},
url={https://openreview.net/forum?id=R73ybUciQF}
}

@inproceedings{
leask2025sparse,
title={Sparse Autoencoders Do Not Find Canonical Units of Analysis},
author={Patrick Leask and Bart Bussmann and Michael T Pearce and Joseph Isaac Bloom and Curt Tigges and Noura Al Moubayed and Lee Sharkey and Neel Nanda},
booktitle={The Thirteenth International Conference on Learning Representations},
year={2025},
url={https://openreview.net/forum?id=9ca9eHNrdH}
}

@misc{cassano2025saemnesia,
      title={SAEmnesia: Erasing Concepts in Diffusion Models with Supervised Sparse Autoencoders}, 
      author={Enrico Cassano and Riccardo Renzulli and Marco Nurisso and Mirko Zaffaroni and Alan Perotti and Marco Grangetto},
      year={2025},
      eprint={2509.21379},
      archivePrefix={arXiv},
      primaryClass={cs.CV},
      url={https://arxiv.org/abs/2509.21379}, 
}

@misc{yang2025alignsae,
      title={AlignSAE: Concept-Aligned Sparse Autoencoders}, 
      author={Minglai Yang and Xinyu Guo and Zhengliang Shi and Jinhe Bi and Steven Bethard and Mihai Surdeanu and Liangming Pan},
      year={2026},
      eprint={2512.02004},
      archivePrefix={arXiv},
      primaryClass={cs.LG},
      url={https://arxiv.org/abs/2512.02004}, 
}

@inproceedings{Hertz2023,
  title={{Prompt-to-Prompt Image Editing with Cross-Attention Control}},
  author={Hertz, Amir and Mokady, Ron and Tenenbaum, Jay and Aberman, Kfir and Pritch, Yael and Cohen-Or, Daniel},
  booktitle={International Conference on Learning Representations (ICLR)},
  year={2023}
}

@inproceedings{Gal2023,
  title={{An Image is Worth One Word: Personalizing Text-to-Image Generation using Textual Inversion}},
  author={Gal, Rinon and Alaluf, Yuval and Atzmon, Yuval and Patashnik, Or and Bermano, Amit H. and Chechik, Gal and Cohen-Or, Daniel},
  booktitle={International Conference on Learning Representations (ICLR)},
  year={2023}
}

@inproceedings{Gandikota2023,
  title={{Erasing Concepts from Diffusion Models}},
  author={Gandikota, Rohit and Materzynska, Joanna and Fiotto-Kaufman, Jaden and Bau, David},
  booktitle={IEEE/CVF International Conference on Computer Vision (ICCV)},
  year={2023}
}

@inproceedings{Kawar2023,
  title={{Imagic: Text-Based Real Image Editing with Diffusion Models}},
  author={Kawar, Bahjat and Zada, Shiran and Lang, Oran and Tov, Omer and Chang, Huiwen and Dekel, Tali and Mosseri, Inbar and Irani, Michal},
  booktitle={IEEE/CVF Conference on Computer Vision and Pattern Recognition (CVPR)},
  year={2023}
}

@inproceedings{ng2011sparseae,
  title={Sparse autoencoder},
  author={Ng, Andrew Y},
  booktitle={CS294A Lecture Notes},
  year={2011},
  note={Stanford University},
  url={https://cs.stanford.edu/~ang/research/nbnotes.pdf}
}

@inproceedings{
suken2023deep,
title={Deep Neural Collapse Is Provably Optimal for the Deep Unconstrained Features Model},
author={Peter S{\'u}ken{\'\i}k and Marco Mondelli and Christoph H Lampert},
booktitle={Thirty-seventh Conference on Neural Information Processing Systems},
year={2023},
url={https://openreview.net/forum?id=v9yC7sSXf3}
}

@inproceedings{alain2016understanding,
  title={Understanding intermediate layers using linear classifier probes},
  author={Alain, Guillaume and Bengio, Yoshua},
  booktitle={International Conference on Learning Representations (ICLR), Workshop Track},
  year={2016}
}

@inproceedings{hewitt2019structural,
  title={A structural probe for finding syntax in word representations},
  author={Hewitt, John and Manning, Christopher D},
  booktitle={Proceedings of the 2019 Conference of the North American Chapter of the Association for Computational Linguistics (NAACL)},
  pages={4129--4138},
  year={2019}
}

@inproceedings{lake2018generalization,
  title={Generalization without systematicity: On the compositional skills of sequence-to-sequence recurrent networks},
  author={Lake, Brenden M and Baroni, Marco},
  booktitle={Proceedings of the 35th International Conference on Machine Learning (ICML)},
  pages={2873--2882},
  year={2018}
}

@inproceedings{10.5555/3692070.3692573,
author = {Esser, Patrick and Kulal, Sumith and Blattmann, Andreas and Entezari, Rahim and M\"{u}ller, Jonas and Saini, Harry and Levi, Yam and Lorenz, Dominik and Sauer, Axel and Boesel, Frederic and Podell, Dustin and Dockhorn, Tim and English, Zion and Rombach, Robin},
title = {Scaling rectified flow transformers for high-resolution image synthesis},
year = {2024},
publisher = {JMLR.org},
booktitle = {Proceedings of the 41st International Conference on Machine Learning},
articleno = {503},
numpages = {28},
location = {Vienna, Austria},
series = {ICML'24}
}

@article{raffel2020t5,
  title={Exploring the Limits of Transfer Learning with a Unified Text-to-Text Transformer},
  author={Raffel, Colin and Shazeer, Noam and Roberts, Adam and Lee, Katherine and Narang, Sharan and Matena, Michael and Zhou, Yanqi and Li, Wei and Liu, Peter J.},
  journal={Journal of Machine Learning Research},
  volume={21},
  number={140},
  pages={1--67},
  year={2020}
}
\bibliographystyle{icml2026}

%%%%%%%%%%%%%%%%%%%%%%%%%%%%%%%%%%%%%%%%%%%%%%%%%%%%%%%%%%%%%%%%%%%%%%%%%%%%%%%
%%%%%%%%%%%%%%%%%%%%%%%%%%%%%%%%%%%%%%%%%%%%%%%%%%%%%%%%%%%%%%%%%%%%%%%%%%%%%%%
% APPENDIX
%%%%%%%%%%%%%%%%%%%%%%%%%%%%%%%%%%%%%%%%%%%%%%%%%%%%%%%%%%%%%%%%%%%%%%%%%%%%%%%
%%%%%%%%%%%%%%%%%%%%%%%%%%%%%%%%%%%%%%%%%%%%%%%%%%%%%%%%%%%%%%%%%%%%%%%%%%%%%%%
\newpage
\appendix
\onecolumn
\section{Additional experimental results}

\begin{figure}[h]
    \centering
    \begin{minipage}{0.30\textwidth}
        \centering
        \includegraphics[width=\textwidth]{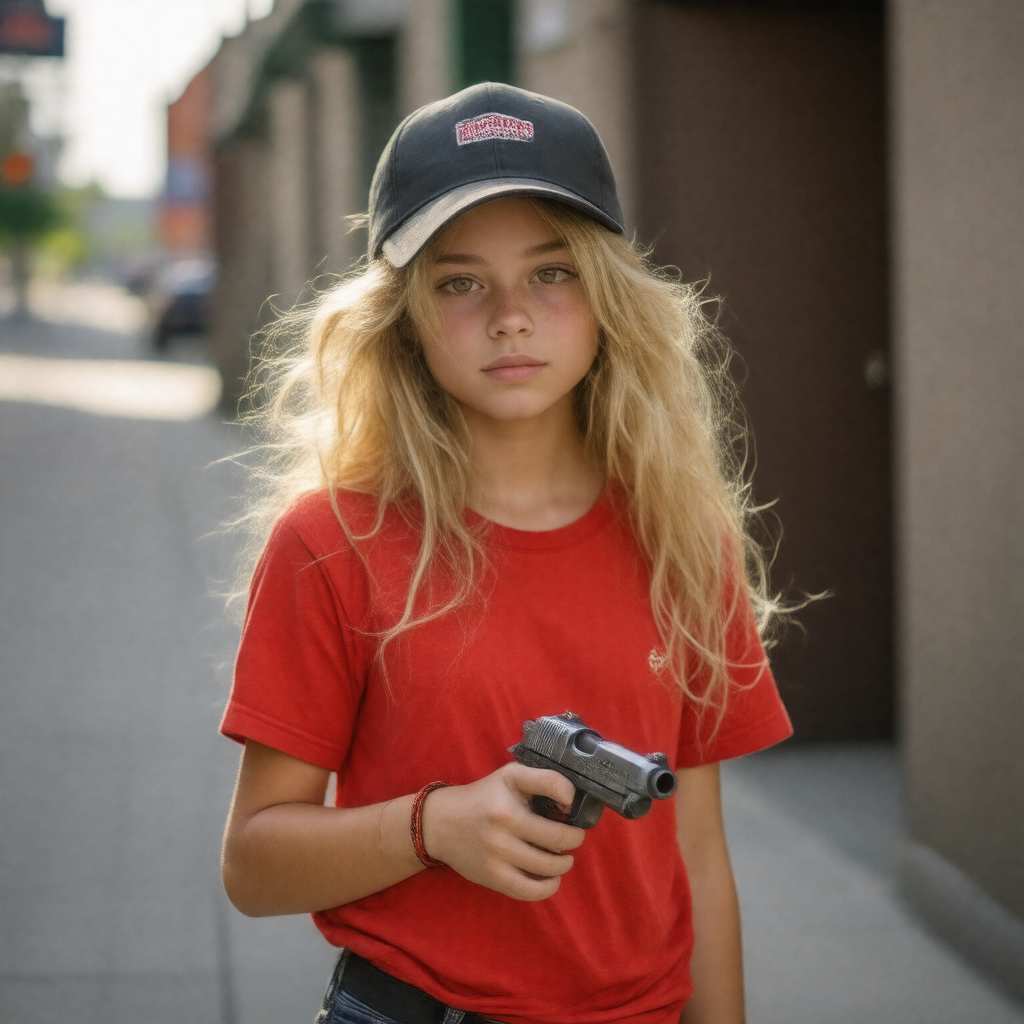}
        \label{fig:image7}
    \end{minipage}
    \begin{minipage}{0.30\textwidth}
        \centering
        \includegraphics[width=\textwidth]{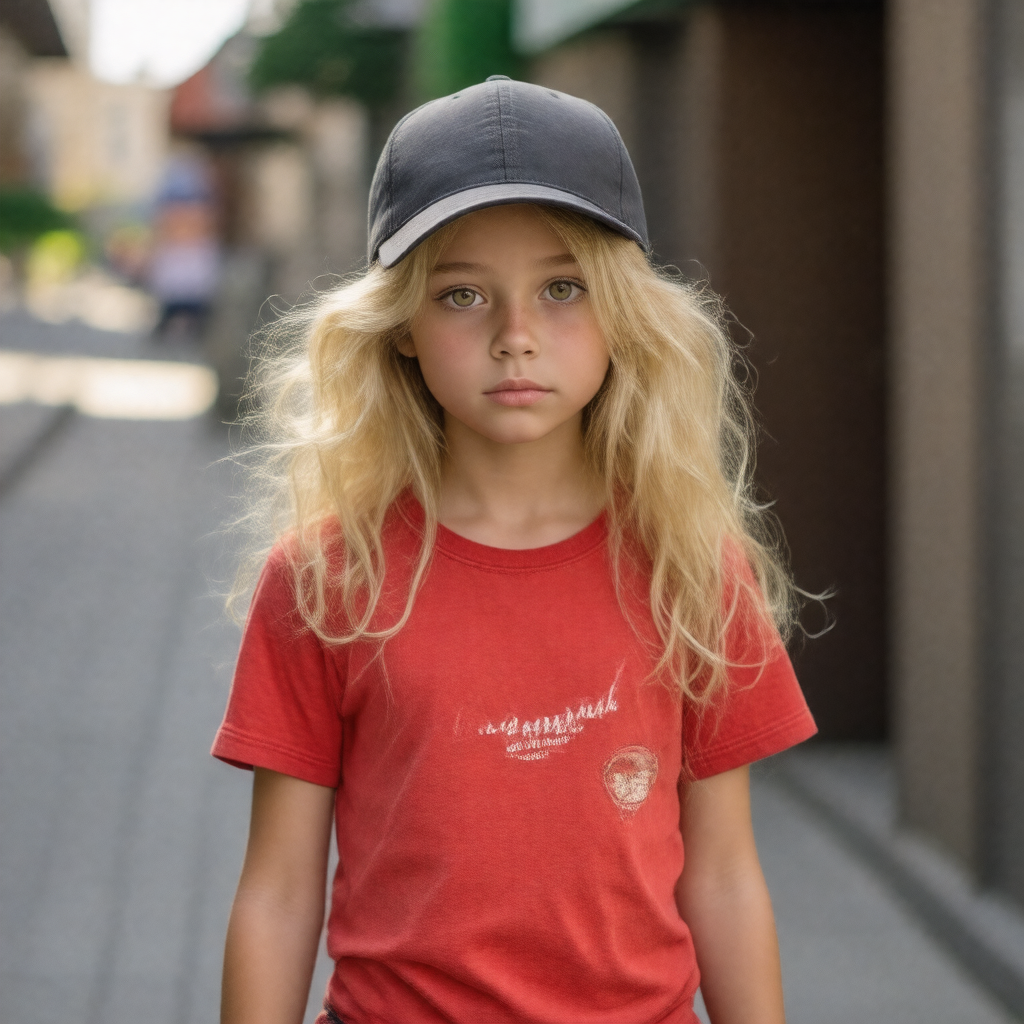}
        \label{fig:image8}
    \end{minipage}
    \\
    \begin{minipage}{0.30\textwidth}
        \centering
        \includegraphics[width=\textwidth]{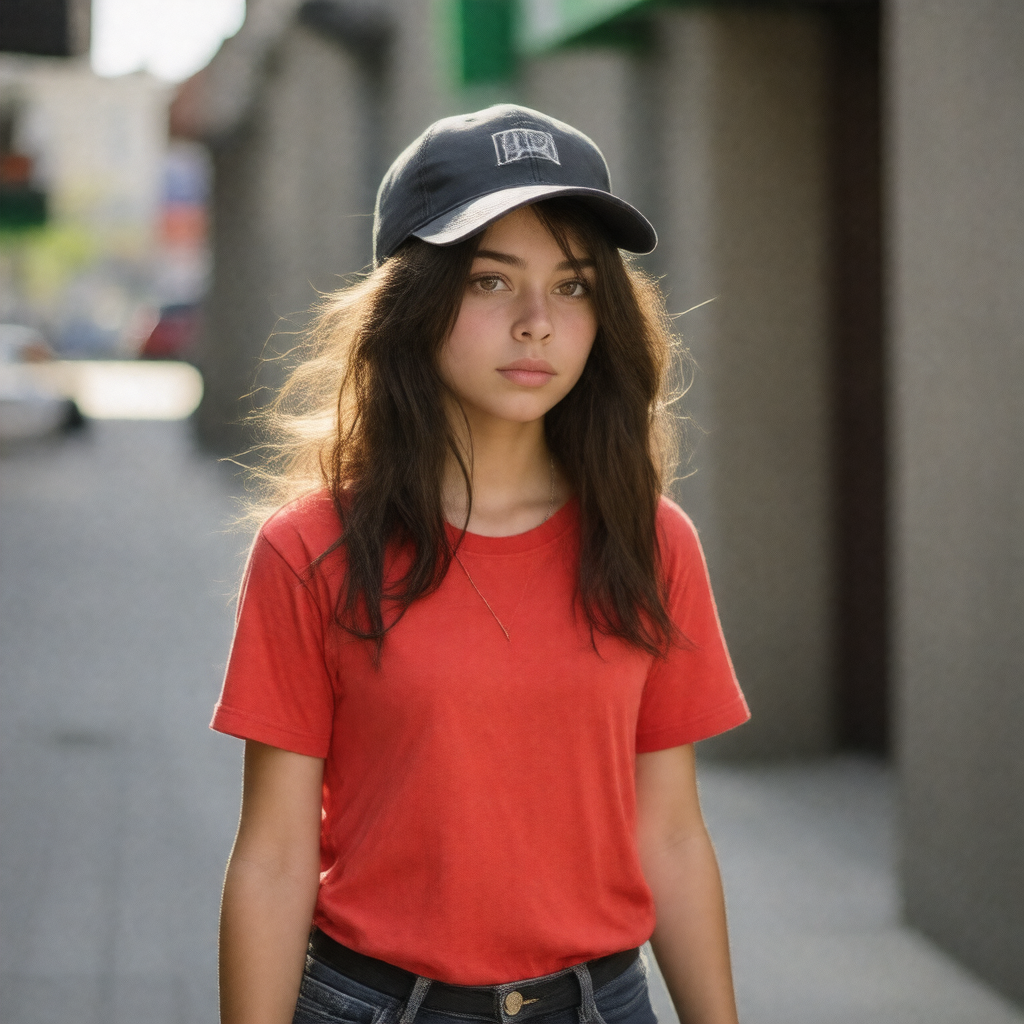}
        \label{fig:image8}
    \end{minipage}
    \begin{minipage}{0.30\textwidth}
        \centering
        \includegraphics[width=\textwidth]{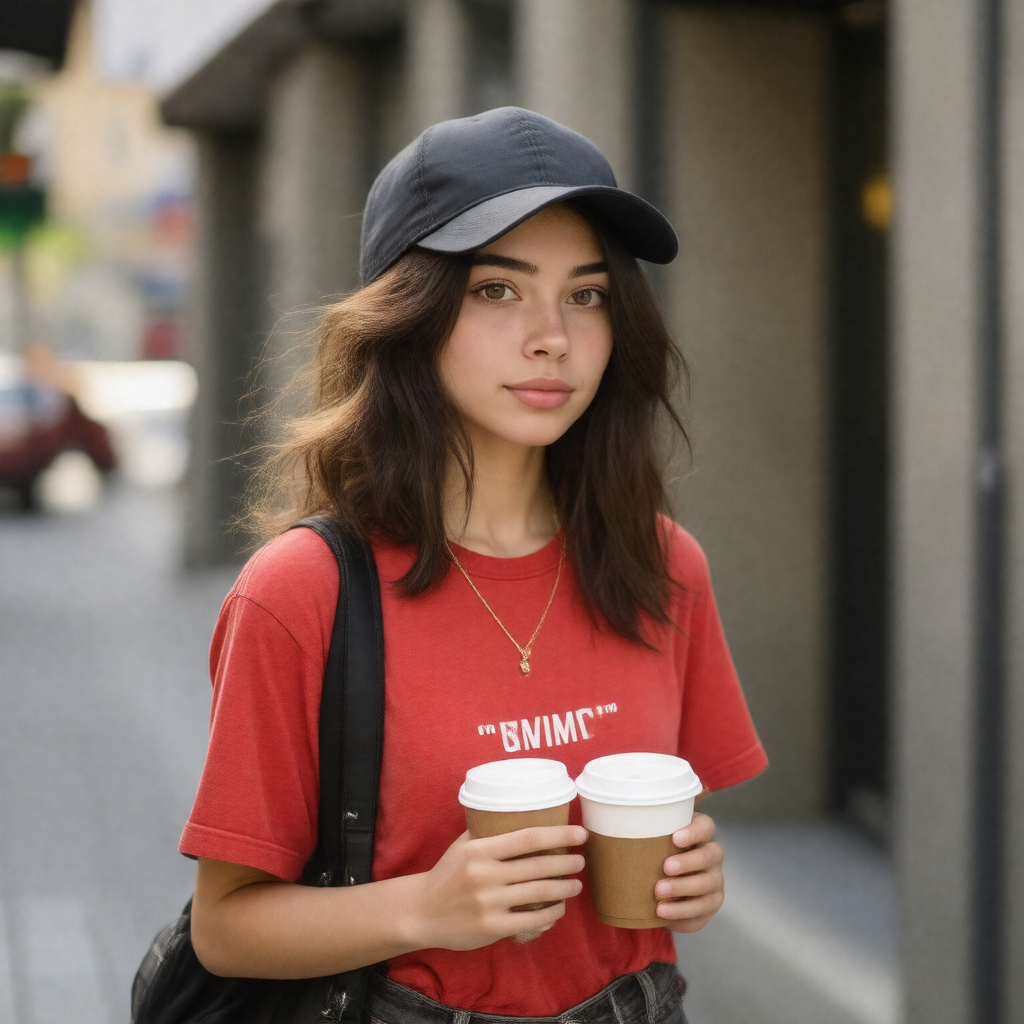}
        \label{fig:image8}
    \end{minipage}
    \caption{Initial prompt: "A blond girl with brown eyes standing in the street, wearing a red t-shirt and a cap, holding a gun, looking in front of her." (top-left), then we applied our transformation in the sparse latent space learnt by our decoder-only SSAE to successively: remove the concept of "holding a gun" (top-right), swap the concepts of "blond hair" and "brune hair" (bottom-left), insert the concept of "holding a coffee" (bottom-right), showcasing compositional generalisation.}
    \label{fig:street}
\end{figure}

\begin{figure}[h]
    \centering
    \begin{minipage}{0.30\textwidth}
        \centering
        \includegraphics[width=\textwidth]{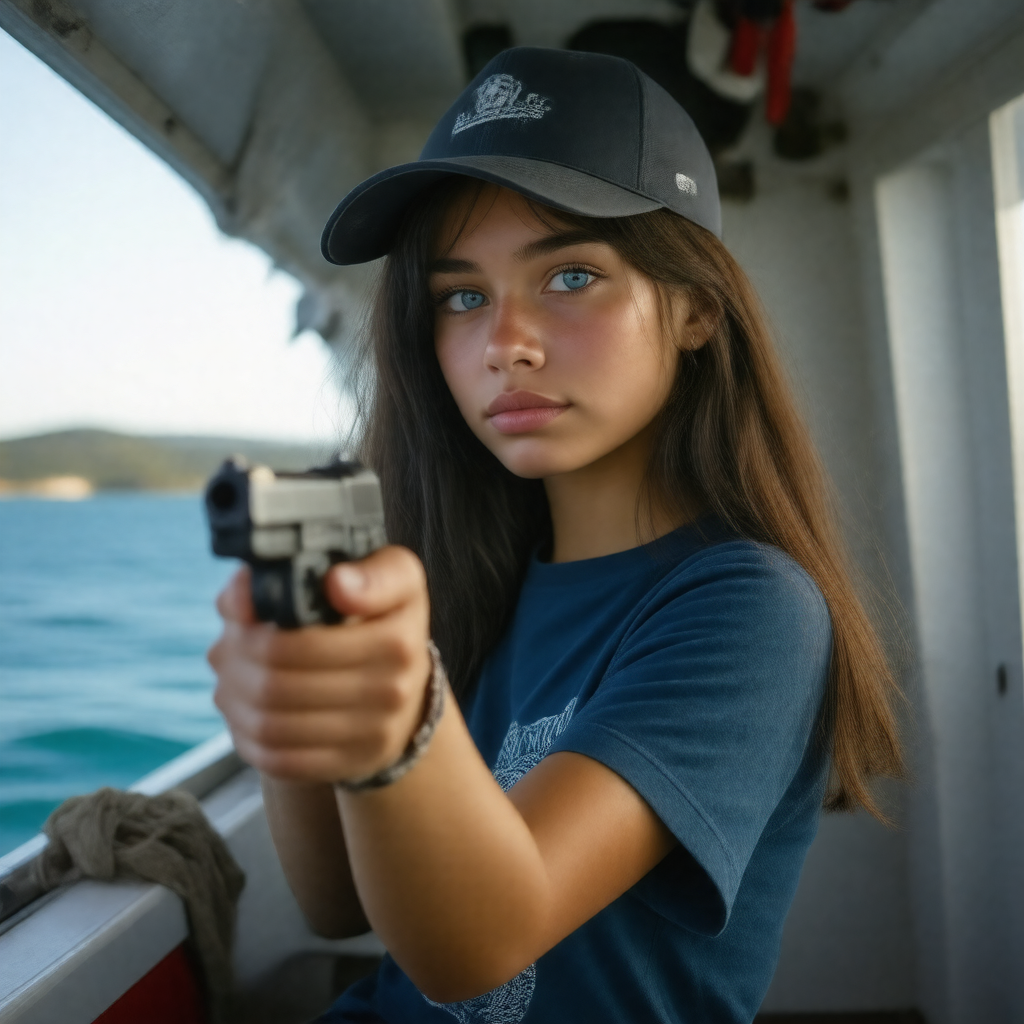}
        \label{fig:image8}
    \end{minipage}
    \begin{minipage}{0.30\textwidth}
        \centering
        \includegraphics[width=\textwidth]{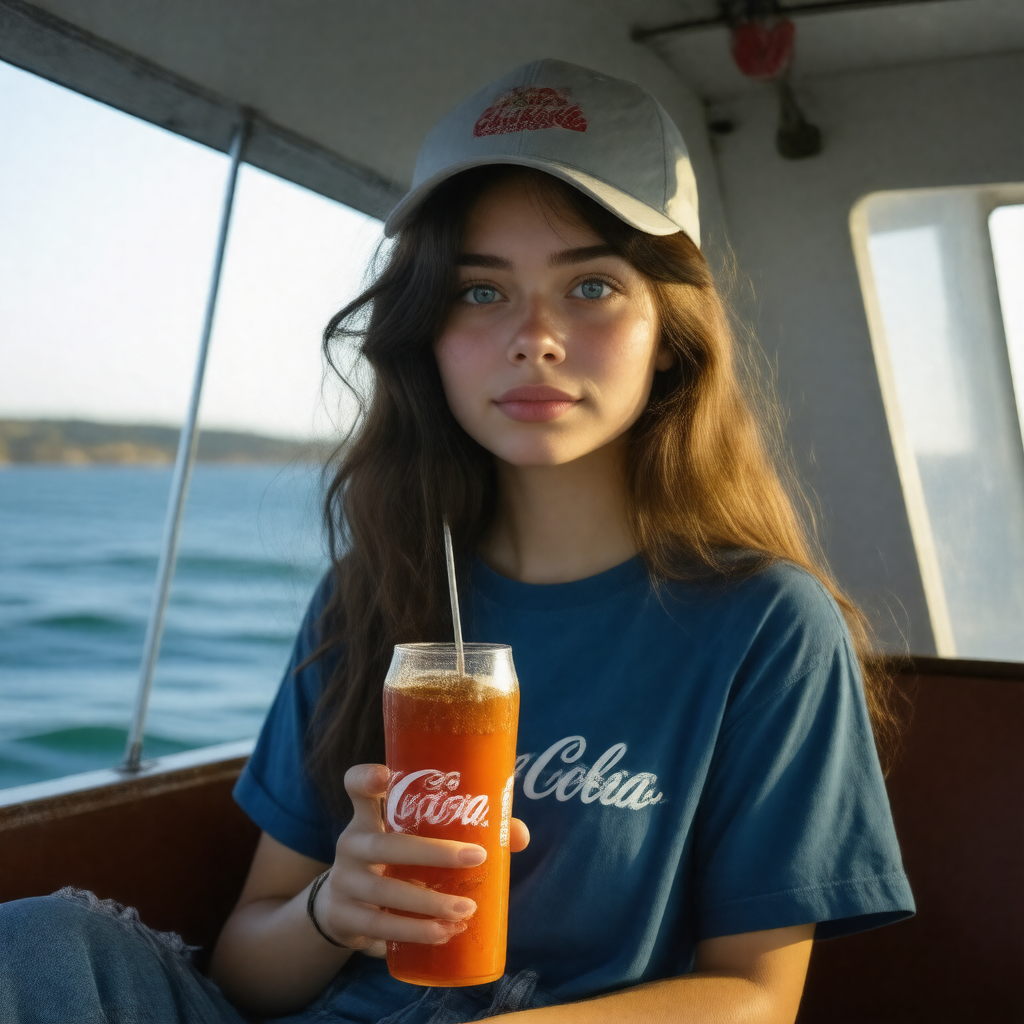}
        \label{fig:image8}
    \end{minipage}
    \\
    \begin{minipage}{0.30\textwidth}
        \centering
        \includegraphics[width=\textwidth]{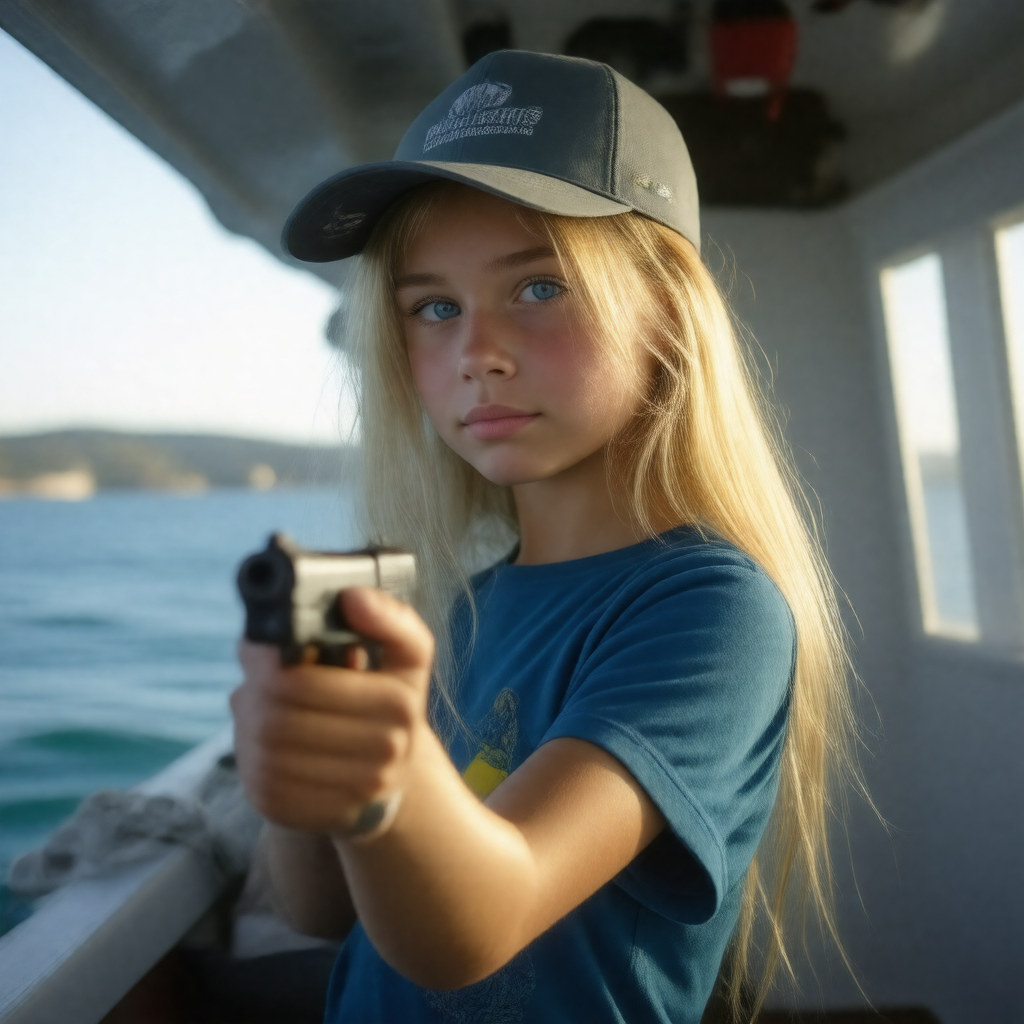}
        \label{fig:image8}
    \end{minipage}
    \begin{minipage}{0.30\textwidth}
        \centering
        \includegraphics[width=\textwidth]{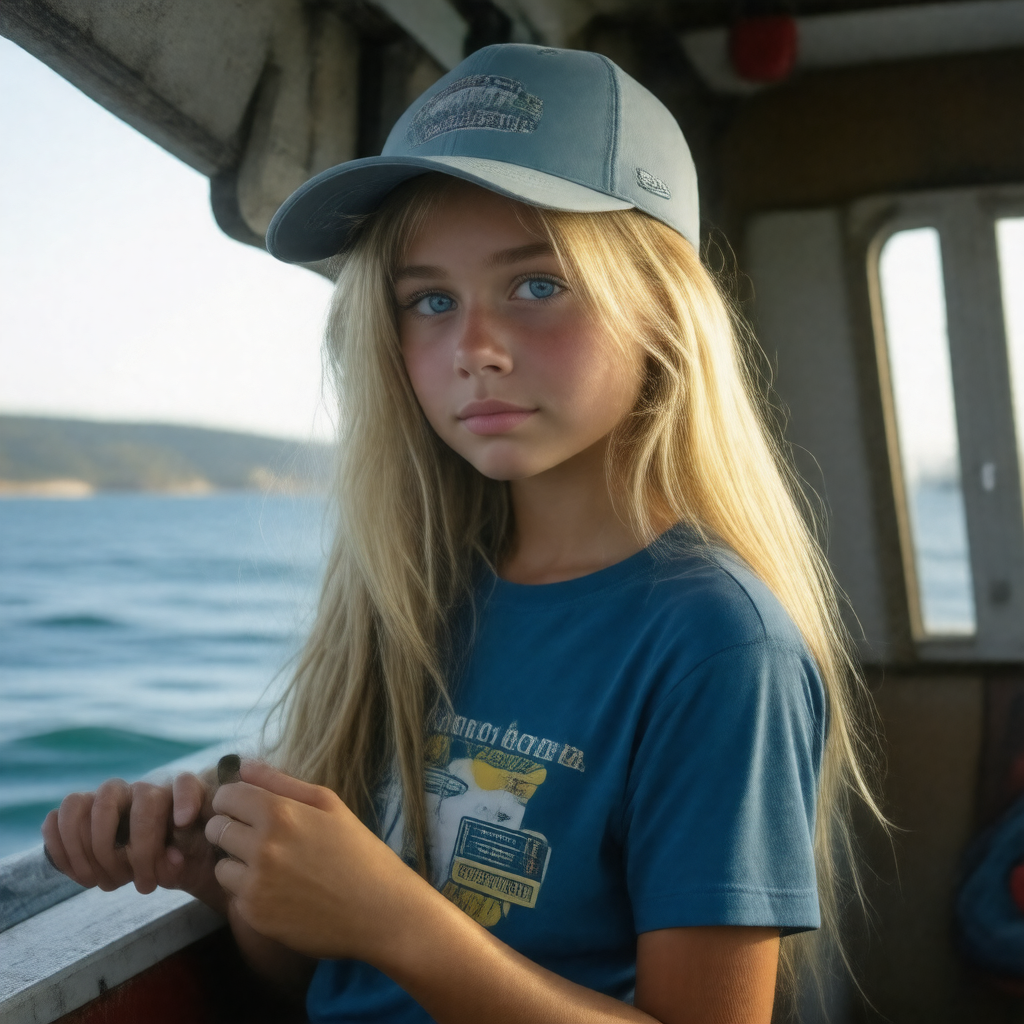}
        \label{fig:image8}
    \end{minipage}
    \caption{Initial prompt: "A brunette girl with blue eyes on a boat, wearing a blue t-shirt and a cap, holding a gun, looking in front of her." (top left), then we applied our transformation in the sparse latent space learnt by our decoder-only SSAE to successively: swap the concept of "holding a gun" with "holding a coca-cola" (from top-left to top-right), swap the concepts of "brune hair" with "blond hair" (from top-left to bottom-left), remove the concept of "holding a gun" (bottom-left to bottom-right), showcasing compositional generalisation.}
    \label{fig:boat}
\end{figure}

%You can have as much text here as you want. The main body must be at most $8$
%pages long. For the final version, one more page can be added. If you want, you
%can use an appendix like this one.

%The $\mathtt{\backslash onecolumn}$ command above can be kept in place if you
%prefer a one-column appendix, or can be removed if you prefer a two-column
%appendix.  Apart from this possible change, the style (font size, spacing,
%margins, page numbering, etc.) should be kept the same as the main body.
%%%%%%%%%%%%%%%%%%%%%%%%%%%%%%%%%%%%%%%%%%%%%%%%%%%%%%%%%%%%%%%%%%%%%%%%%%%%%%%
%%%%%%%%%%%%%%%%%%%%%%%%%%%%%%%%%%%%%%%%%%%%%%%%%%%%%%%%%%%%%%%%%%%%%%%%%%%%%%%

\end{document}